\documentclass{article}

\usepackage{microtype}
\usepackage{graphicx}
\usepackage{booktabs} 

\usepackage{hyperref}

\usepackage[accepted]{icml2023}

\usepackage{amsmath}
\usepackage{amssymb}
\usepackage{mathtools}
\usepackage{amsthm}

\usepackage[capitalize,nameinlink]{cleveref}

\theoremstyle{plain}
\newtheorem{theorem}{Theorem}[section]

\theoremstyle{definition}

\newtheorem{assumption}[theorem]{Assumption}
\theoremstyle{remark}

\usepackage[textsize=tiny]{todonotes}

\usepackage{array}
\usepackage{tikz}
\usetikzlibrary{calc}
\usetikzlibrary{plotmarks}
\usetikzlibrary{shapes.geometric}
\usepackage{xstring}  
\usepackage[skip=6pt]{caption}
\usepackage{subcaption}
\usepackage{enumitem}   
\setlist{leftmargin=3.5mm}
\usepackage{wrapfig}
\usepackage{pgfplots}
\pgfplotsset{compat=1.18} 
\usepackage{bbm}
\usepackage{algorithm}
\usepackage{algorithmic}

\newcommand{\PreserveBackslash}[1]{\let\temp=\\#1\let\\=\temp}
\newcolumntype{C}[1]{>{\PreserveBackslash\centering}p{#1}}
\newcolumntype{R}[1]{>{\PreserveBackslash\raggedleft}p{#1}}
\newcolumntype{L}[1]{>{\PreserveBackslash\raggedright}p{#1}}

\newcommand{\observation}{%
    \begin{tikzpicture}[inner sep=0pt, baseline=(base)]%
    \node (base) at (0,-.5ex) {};
    \fill [draw=orange, fill=white] (0,0) circle (2pt) node [] {};
    \end{tikzpicture}%
}
\newcommand{\treatment}{%
    \begin{tikzpicture}[inner sep=0pt, baseline=(base), sibling distance=80pt,square/.style={rectangle,draw}]%
    \node (base) at (0,-.5ex) {};
    \draw node[square, draw=orange, fill=orange, inner sep=0pt, minimum width=5pt, minimum height=5pt] (d) at (0,0) {};
    \end{tikzpicture}%
}
\newcommand{\inputarrow}{%
    \begin{tikzpicture}[inner sep=0pt, baseline=(base)]
    \draw[->, densely dotted] (-0.2, 0) -- (0.2, 0);
    \end{tikzpicture}%
}
\newcommand{\forwardarrow}{%
    \begin{tikzpicture}[inner sep=0pt, baseline=(base)]
    \draw[->] (-0.2, 0) -- (0.2, 0);
    \end{tikzpicture}%
}
\newcommand{\backproparrow}{%
    \begin{tikzpicture}[inner sep=0pt, baseline=(base)]
    \draw[stealth-, dotted] (-0.2, 0) -- (0.2, 0);
    \end{tikzpicture}%
}
\newcommand{\nobackproparrow}{%
    \begin{tikzpicture}[inner sep=0pt, baseline=(base)]
    \draw[->, dashed] (-0.2, 0) -- (0.2, 0);
    \end{tikzpicture}%
}
\tikzset{declare function={f(\x) = ((sin(0.5 * deg(\x)) + cos(2 * deg(\x)) - 0.5*cos(3 * deg(\x)) + (\x/5 - 3)^2) - x) / 10 + 0.1;}}
\newcommand{\VerticalLine}[2][]{
\addplot[#1] coordinates {(#2,0) (#2, {f(#2)})};}

\icmltitlerunning{Accounting For Informative Sampling When Learning to Forecast Treatment Outcomes Over Time}

\begin{document}

\twocolumn[
\icmltitle{Accounting For Informative Sampling When Learning to Forecast Treatment Outcomes Over Time}



\icmlsetsymbol{equal}{*}

\begin{icmlauthorlist}
\icmlauthor{Toon Vanderschueren}{equal,kul,ant}
\icmlauthor{Alicia Curth}{equal,cam}
\icmlauthor{Wouter Verbeke}{kul}
\icmlauthor{Mihaela van der Schaar}{cam,ati}
\end{icmlauthorlist}

\icmlaffiliation{kul}{Decision Sciences and Information Management, KU Leuven}
\icmlaffiliation{ant}{Applied Mathematics, University of Antwerp}
\icmlaffiliation{cam}{DAMTP, University of Cambridge}
\icmlaffiliation{ati}{The Alan Turing Institute}

\icmlcorrespondingauthor{Toon Vanderschueren}{toon.vanderschueren@kuleuven.be}
\icmlcorrespondingauthor{Alicia Curth}{amc253@cam.ac.uk}

\icmlkeywords{Treatment effects, Treatment effects over time, Irregular sampling, Informative sampling}

\vskip 0.3in
]



\printAffiliationsAndNotice{\icmlEqualContribution} 

\begin{abstract}
Machine learning (ML) holds great potential for accurately forecasting treatment outcomes over time, which could ultimately enable the adoption of more individualized treatment strategies in many practical applications. However, a significant challenge that has been largely overlooked by the ML literature on this topic is the presence of \textit{informative sampling} in observational data. When instances are observed irregularly over time, sampling times are typically not random, but rather informative--depending on the instance's characteristics, past outcomes, and administered treatments. In this work, we formalize informative sampling as a covariate shift problem and show that it can prohibit accurate estimation of treatment outcomes if not properly accounted for. To overcome this challenge, we present a general framework for learning treatment outcomes in the presence of informative sampling using inverse intensity-weighting, and propose a novel method, TESAR-CDE, that instantiates this framework using Neural CDEs. Using a simulation environment based on a clinical use case, we demonstrate the effectiveness of our approach in learning under informative sampling.
\end{abstract}

\section{Introduction}

Due to its importance in applications ranging from economics to healthcare and marketing, the problem of estimating personalized causal effects of actions -- e.g., treatments, interventions, or policies -- has received wide attention in the recent machine learning (ML) literature \cite{curth2021nonparametric}. Effectively using real data for estimating such effects requires dealing with unique challenges arising from its observational nature. Therefore, the recent ML literature on treatment effect estimation has paid great attention to solving methodological issues arising due to treatment assignment biases in static \citep{shalit2017estimating} and longitudinal settings \citep{bica2019estimating}. This paper focuses on another challenge that has been largely overlooked by the ML literature on treatment effect estimation, despite its relevance and prevalence in practice: the problem of \textit{informative sampling}, sometimes also called \textit{informed presence bias} \cite{goldstein2016controlling}. 
That is, in observational data, the timing at which an observation was made is often not random, but rather indicative of some underlying information relevant to the estimation problem of interest.

In electronic health records, for example, patients are typically not recorded \textit{randomly} over time, but \textit{informatively} \cite{lin2004analysis}: observations are only recorded at irregular visits to a health care provider, with visit times typically depending on the patient's past and present characteristics, evolving health state, and administered treatments. The resulting sampling mechanism is inherently intertwined with the patient's observed outcomes and treatments, with more check-ups being scheduled for patients in critical condition or to follow up after a treatment. Throughout this work, we will refer to examples from health care due to their societal relevance and intuitive appeal, but the problem of informative sampling appears in a wide variety of other domains, such as policy design \citep{lin2004analysis}, epidemiology \citep{del2015sequential}, economics \citep{clithero2018response}, or maintenance \citep{vanderschueren2023optimizing}. 

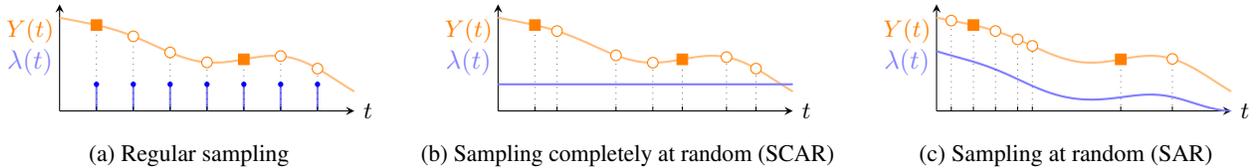
\begin{figure*}[t]
\begin{subfigure}[b]{0.32\textwidth}
\centering
\begin{tikzpicture}
\begin{axis}[every axis plot post/.append style={}, 
    axis x line=bottom, 
    axis y line=left, 
    xmin=0, xmax=4,
    ymin=-0, ymax=1.2,
    width=5.5cm, height=3cm,
    xmajorticks=false,
    ymajorticks=false,
    xlabel={\small $t$},
    ylabel={\small \textcolor{orange}{$Y(t)$} \\ \textcolor{blue!50}{$\lambda(t)$}},
    x label style={at={(axis description cs:1,0)},anchor=west},
    y label style={at={(axis description cs:-0.1,.25)},rotate=-90,anchor=south, align=center}
    ]    
    \addplot [thick, orange!50, domain=0:4, samples=50, smooth] {f(x)};
    \draw [blue!50, thick] (0.5, 0.0) -- (0.5, 0.3);
    \draw [blue!50, thick] (1, 0.0) -- (1, 0.3);
    \draw [blue!50, thick] (1.5, 0.0) -- (1.5, 0.3);
    \draw [blue!50, thick] (2, 0.0) -- (2, 0.3);
    \draw [blue!50, thick] (2.5, 0.0) -- (2.5, 0.3);
    \draw [blue!50, thick] (3, 0.0) -- (3, 0.3);
    \draw [blue!50, thick] (3.5, 0.0) -- (3.5, 0.3);
    \fill [blue] (0.5, 0.3) circle (1pt) node [] {};
    \fill [blue] (1, 0.3) circle (1pt) node [] {};
    \fill [blue] (1.5, 0.3) circle (1pt) node [] {};;
    \fill [blue] (2, 0.3) circle (1pt) node [] {};
    \fill [blue] (2.5, 0.3) circle (1pt) node [] {};;
    \fill [blue] (3, 0.3) circle (1pt) node [] {};
    \fill [blue] (3.5, 0.3) circle (1pt) node [] {};
    \foreach \c in {0.5, 1, 1.5, 2, 2.5, 3, 3.5}{
        \VerticalLine[very thin, dotted]{\c};
    }
    \draw [thin, solid] (0.5,-0.05) -- (0.5, 0.05);
    \draw [thin, solid] (1,-0.05) -- (1, 0.05);
    \draw [thin, solid] (1.5,-0.05) -- (1.5, 0.05);
    \draw [thin, solid] (2,-0.05) -- (2, 0.05);
    \draw [thin, solid] (2.5,-0.05) -- (2.5, 0.05);
    \draw [thin, solid] (3,-0.05) -- (3, 0.05);
    \draw [thin, solid] (3.5,-0.05) -- (3.5, 0.05);
    \addplot[only marks, samples at={0.5,2.5}, mark=square*, mark options={fill=orange,scale=1, solid, color=orange}]{f(x)};
    \addplot[only marks, samples at={1, 1.5, 2, 3, 3.5}, mark=*, mark options={color=orange, fill=white, scale=1, solid}]{f(x)};
\end{axis}
\end{tikzpicture}
\caption{{Regular sampling}}
\label{sfig:regular_sampling}
\end{subfigure}
\hfill
\begin{subfigure}[b]{0.32\textwidth}
\centering
\begin{tikzpicture}
\begin{axis}[every axis plot post/.append style={}, 
    axis x line=bottom, 
    axis y line=left, 
    xmin=0, xmax=4,
    ymin=0, ymax=1.2,
    width=5.5cm, height=3cm,
    xmajorticks=false,
    ymajorticks=false,
    xlabel={\small $t$},
    ylabel={\small \textcolor{orange}{$Y(t)$} \\ \textcolor{blue!50}{$\lambda(t)$}},
    x label style={at={(axis description cs:1,0)},anchor=west},
    y label style={at={(axis description cs:-0.1,.25)},rotate=-90,anchor=south, align=center}
    ]    
    \addplot [thick, orange!50, domain=0:4, samples=50, smooth] {f(x)};
    \addplot [thick, blue!50, domain=0:4, samples=50, smooth] {0.3};
    \foreach \c in {0.5, 0.8, 1.6, 2.1, 2.5, 3.1, 3.5}{
        \VerticalLine[very thin, dotted]{\c};
    }
    \draw [thin, solid] (0.5,-0.05) -- (0.5, 0.05);
    \draw [thin, solid] (0.8,-0.05) -- (0.8, 0.05);
    \draw [thin, solid] (1.6,-0.05) -- (1.6, 0.05);
    \draw [thin, solid] (2.1,-0.05) -- (2.1, 0.05);
    \draw [thin, solid] (2.5,-0.05) -- (2.5, 0.05);
    \draw [thin, solid] (3.1,-0.05) -- (3.1, 0.05);
    \draw [thin, solid] (3.5,-0.05) -- (3.5, 0.05);
    \addplot[only marks, samples at={0.5,2.5}, mark=square*, mark options={fill=orange,scale=1, solid, color=orange}]{f(x)};
    \addplot[only marks, samples at={0.8, 1.6, 2.1, 3.1, 3.5}, mark=*, mark options={color=orange, fill=white, scale=1, solid}]{f(x)};
\end{axis}
\end{tikzpicture}
\caption{{Sampling completely at random (SCAR)}}
\label{sfig:scar}
\end{subfigure}
\hfill
\begin{subfigure}[b]{0.32\textwidth}
\centering
\begin{tikzpicture}
\begin{axis}[every axis plot post/.append style={}, 
    axis x line=bottom, 
    axis y line=left, 
    xmin=0, xmax=4,
    ymin=0, ymax=1.2,
    width=5.5cm, height=3cm,
    xmajorticks=false,
    ymajorticks=false,
    xlabel={\small $t$},
    ylabel={\small \textcolor{orange}{$Y(t)$} \\ \textcolor{blue!50}{$\lambda(t)$}},
    x label style={at={(axis description cs:1,0)},anchor=west},
    y label style={at={(axis description cs:-0.1,.25)},rotate=-90,anchor=south, align=center}
    ]    
    \addplot [thick, orange!50, domain=0:4, samples=50, smooth] {f(x)};
    \addplot [thick, blue!50, domain=0:4, samples=50, smooth] {0.8*f(x)*f(x) - 0.2*f(x)};
    \foreach \c in {0.2, 0.5, 0.8, 1.1, 1.3, 2.5, 3.2}{
        \VerticalLine[very thin, dotted]{\c};
    }
    \draw [thin, solid] (0.2,-0.05) -- (0.2, 0.05);
    \draw [thin, solid] (0.5,-0.05) -- (0.5, 0.05);
    \draw [thin, solid] (0.8,-0.05) -- (0.8, 0.05);
    \draw [thin, solid] (1.1,-0.05) -- (1.1, 0.05);
    \draw [thin, solid] (1.3,-0.05) -- (1.3, 0.05);
    \draw [thin, solid] (2.5,-0.05) -- (2.5, 0.05);
    \draw [thin, solid] (3.2,-0.05) -- (3.2, 0.05);
    \addplot[only marks, samples at={0.5,2.5}, mark=square*, mark options={fill=orange,scale=1, solid, color=orange}]{f(x)};
    \addplot[only marks, samples at={0.2, 0.8, 1.1, 1.3, 3.2}, mark=*, mark options={color=orange, fill=white, scale=1, solid}]{f(x)};
\end{axis}
\end{tikzpicture}
\caption{{Sampling at random (SAR)}}
\label{sfig:sar}
\end{subfigure}
\caption{\textbf{Problem illustration: sampling mechanisms.} We show an instance's latent trajectory (\textcolor{orange}{$Y(t)$}) and sampling intensity (\textcolor{blue!50}{$\lambda(t)$}) over time $t$, along with administered treatments (\treatment) and observations (\observation) resulting from different sampling mechanisms. \textbf{(a) Regular.} Samples are obtained at regular intervals over time. \textbf{(b) SCAR.} Samples are irregular, drawn at completely random intervals over time. \textbf{(c) SAR.} Sampling times are \textit{irregular, but not completely random}: e.g., there might be more samples when the outcome is large. We refer to the dependence of the sampling intensity on an instance's covariates, treatments, and/or outcomes as informative sampling. Whereas existing work in the ML literature assumes regular sampling or SCAR, this work is, to the best of our knowledge, the first to consider learning to forecast treatment outcomes given SAR.}
\vspace{-2pt}
\hrulefill
\vspace{-12pt}
\label{fig:problem_overview}
\end{figure*}

Informative sampling poses an important challenge as it can bias estimates of causal effects when not accounted for \citep{robins1995analysis, lin2004analysis, mcculloch2016biased}. Intuitively, informative sampling leads to relatively more measurements of abnormal values and fewer measurements of normal values and, therefore, selection bias in the data \citep{liu2008analysis, gasparini2020mixed}. Standard statistical methods can estimate causal effects in the presence of informative sampling given a well-specified model of the sampling mechanism  \citep{hernan2009observation}. However, existing approaches for modeling the sampling mechanism from the (bio)statistics literature assume a certain parametric form or latent variable(s), which might not match the actual data-generating process. Moreover, \citet{farzanfar2017longitudinal}'s survey on longitudinal healthcare research finds that these methods are rarely used in practice, leaving potential bias largely unaddressed. Therefore, this work examines the use of flexible ML methods for this task and investigates the unique methodological challenges arising therein.

\textbf{Related work.\footnote{We discuss the related work more extensively in \cref{sec:appendix_related_work}.}} Since the initial seminal ML work on heterogeneous treatment effect estimation considering binary treatments and static data \cite{johansson2016learning, shalit2017estimating}, this literature has grown rapidly both by making methodological refinements in the original setting \cite{hassanpour2019learning, curth2021inductive} and by considering new settings, such as continuous treatments \cite{bica2019estimating} or survival outcomes \cite{curth2021survite}. Recent extensions have specifically explored using ML methods for estimating treatment effects \textit{over time}, such as RNNs \citep{lim2018rmsn, bica2019estimating, li2021gnet, berrevoets2021disentangled}, transformers \citep{melnychuk2022causaltransf}, and Neural ODEs \citep{gwak2020neural,de2022predicting} or Neural CDEs \citep{seedat2022continuous}.

Most existing ML work on causal inference in a temporal setting has, to the best of our knowledge, implicitly relied on strict assumptions regarding the sampling mechanism. The majority assumes regular and uninformative sampling times (\cref{sfig:regular_sampling}). Only very recent work relying on neural differential equations to model the effects of treatment in continuous time  \cite{gwak2020neural,seedat2022continuous,de2022predicting} \textit{allows} for observations to be irregular (\cref{sfig:scar}), but \textit{does not consider or account for potential bias} resulting from sampling times being informative rather than completely random, which is the focus of this work (\cref{sfig:sar}). This stands in stark contrast to the close attention paid in the treatment effect estimation literature to other sources of \textit{covariate shift} arising in observational data, e.g., due to static treatment assignment \cite{johansson2016learning}, treatment assignment over time \cite{bica2019estimating}, censoring \cite{curth2021survite} or competing events \cite{curth2023understanding}. In this spirit, we find it important to study when and how the informativeness of sampling acts as an additional source of covariate shift in this setting.


\textbf{Contributions.} Despite the rapid recent expansion of the ML literature on estimating treatment effects, we believe that there is still a fundamental lack of understanding, or even formalization, of the challenges arising due to one of the most fundamental features of observational data: \textit{sampling, the act of observation itself, can be inherently informative}. Therefore, we focus on understanding and analyzing the challenges that arise from informative sampling and propose strategies to alleviate bias arising in this context. In doing so, we make three contributions: 
\textbf{(1)} We formalize the problem of forecasting treatment outcomes under informative sampling as a machine learning problem and characterize the key challenges arising therein as a consequence of covariate shift induced by informative sampling.
\textbf{(2)}  We present a general strategy for tackling this challenge and propose a novel method for learning under informative sampling, TESAR-CDE, that instantiates this framework using Neural CDEs.
\textbf{(3)} We design a simulation environment based on a clinical use case to study the effect and different drivers of informative sampling and use it to empirically demonstrate that our proposed method is able to correct for the resulting bias that existing methods can suffer from.

\section{Problem Formalization: Data Structure and Informative Sampling Mechanism}\label{sec:formaliz}

This section describes and formalizes the problem of forecasting treatment outcomes in the presence of informative sampling. We investigate the assumptions required for and challenges inherent to tackling this problem in \cref{sec:capo}. We build on the exposition in \citet{lin2004analysis}, who study longitudinal \textit{outcome prediction} in the presence of informative sampling but do not explicitly consider estimating treatment effects, and build on ideas from \citet{lok2008statistical, seedat2022continuous} who study forecasting treatment outcomes in continuous time but do not consider informative sampling.
{\parfillskip=0pt \emergencystretch=.5\textwidth \par}

\subsection{Problem structure: Complete versus observed data}
\textbf{Underlying complete data structure.} We consider data collected over a period $[0, T]$ in which instances are characterized by a $d$-dimensional covariate path $X: [0, T] \rightarrow \mathbb{R}^d$ and a treatment path $A: [0, T] \rightarrow  \{0, 1\}$ -- which jumps to $1$ only at time steps $t$ when treatment is administered\footnote{In this exposition, we assume that $A(t)=1$ only at single time-steps where treatment is administered; for treatments that are administered over a time-period $[t_1, t_2]$ one could instead define a counting process that jumps whenever treatment status is \textit{changed} as in \citet{lok2008statistical, seedat2022continuous}. Further, as noted in \citet{seedat2022continuous}, this definition can be generalized to multiple treatments by assuming $A$ to be a multivariate process.} -- both of which possibly modulate an outcome process of interest $Y: [0, T] \rightarrow \mathbb{R}$. While we only observe the outcome $Y$ associated with the treatment path $A$ that was actually administered (sometimes also referred to as the \textit{factual outcome}), we assume that any instance is characterised by a possibly infinite number of \textit{potential outcomes} $Y_a:[0, T] \rightarrow \mathbb{R}$ associated with other feasible treatment paths $a$.

\textbf{Observed data structure.} Paths $X$, $A$ and $Y$ are only sampled (observed) at possibly irregular and discrete time-points, such as scheduled check-ups or unscheduled appointments. Therefore, we additionally define a counting process $N: [0,T] \rightarrow \mathbb{N}_0$ recording the number of observations made by time $t$. This process jumps whenever a new observation is sampled\footnote{For ease of exposition, we assume that whenever an instance is observed at $t$, we record all of $X(t)$, $A(t)$ and $Y(t)$. Nevertheless, it would be possible to relax this by instead introducing separate counting processes for each variable or component thereof.}, so that $dN(t)=1$ if an instance is sampled at time $t$ and $dN(t)=0$ otherwise, where $dN(t) = N(t) - \lim_{s \uparrow t} N(s)$. As in \citet{lin2004analysis}, for a variable $V(t)$, let $V^{o}(t)=V(\max s: 0 \leq t, dN(s)=1)$ denote its most recent observation by time $t$, $ \bar{V}^{o}(t)=\{V^{o}(s): 0 \leq s \leq t\}$ its observed history by time $t$ and $\bar{V}(t)=\{V(s): 0 \leq s \leq t\}$ its full (yet possibly not observed) history by time $t$. Further, let $\bar{V}^{o}(t^-)$ and $\bar{V}(t^-)$ denote the same histories where the upper limit does \textit{not} include $t$. Then, for a study following $n$ instances until time $T$, we observe a dataset $\mathcal{D}=\{\mathcal{O}_i\}^n_{i=1}$ consisting of $n$ i.i.d. copies of $\mathcal{O}=\mathcal{F}^{o}(T)$ where $
 \mathcal{F}^{o}(T)=(\bar{X}^{o}(T), \bar{N}(T),  \bar{A}^{o}(T), \bar{Y}^{o}(T))$.

\subsection{Distinguishing between different sampling patterns} In order to define what distinguishes the \textit{informativeness} of different sampling patterns, we first need to introduce a conditional \textit{intensity} $\lambda(t)$ which governs the observation process $N(t)$. In its most general form, adopting the notation of \citet{lin2004analysis}, this can be defined through
\begin{equation}
  \mathbb{P}(dN(t)\!=\!1|  \bar{X}(T), \bar{N}(T), \bar{A}(T), \bar{Y}(T)) = \lambda(t)dt,
\end{equation}
For notational convenience, we omit conditioning in $\lambda(t)$.

Using this definition, we can differentiate between different sampling mechanisms \citep{pullenayegum2016longitudinal}, giving rise to a classification similar to missingness mechanisms in static data \cite{rubin1976inference}. This categorization is based on the causal role of the instance history in relation to the observation intensity (\cref{fig:problem_overview} shows a graphical overview):

\begin{itemize}[itemsep=-1pt, topsep=-4pt]
\setlength\itemsep{0pt}
    \item \textbf{Regular sampling ({\normalfont \cref{sfig:regular_sampling}}).}
    Instances are observed at $K$ regular (pre-determined) timesteps $\mathcal{T} = \{t_1, \ldots, t_K\}$, so that $\lambda(t)dt=1 \text{ if } t \in \mathcal{T} \text{ else } \lambda(t)dt=0$. Most existing related work \citep[e.g.][]{lim2018rmsn, bica2019estimating, melnychuk2022causaltransf} implicitly relies on this assumption.
    
    \item 
    \textbf{Sampling completely at random (SCAR{; \normalfont \cref{sfig:scar}}).} Instances are observed at completely random time-steps, with the intensity independent of all variables: $ \lambda(t)dt\!=\! \mathbb{P}(dN(t)\!=\!1|  \bar{X}(T), \bar{N}(T),  \bar{A}(T),  \bar{Y}(T))\!=\! \mathbb{P}(dN(t)\!=\!1)$. Recent work on treatment effect estimation from irregularly sampled data using neural differential equations \citep{gwak2020neural, seedat2022continuous, de2022predicting} is explicitly only equipped to handle this scenario. 
    
    \item 
    \textbf{Sampling at random (SAR{; \normalfont \cref{sfig:sar}}).} Being observed at time $t$ is independent of the (up until then unknown) outcome at time $t$ given the observed history up to time $t$: 
    \begin{equation*}
    \begin{split}
        \lambda(t)dt=\mathbb{P}(dN(t)\!=\!1|  \bar{X}(T),  \bar{N}(T), \bar{A}(T),  \bar{Y}(T)) \\= \mathbb{P}(dN(t)\!=\!1| \bar{X}^{o}(t), \bar{N}(t^-),\bar{A}^{o}(t^-),  \bar{Y}^{o}(t^-))
    \end{split}
    \end{equation*}   
    This work investigates the challenges of learning given SAR, which is considerably weaker than SCAR: it allows, for example,  for patients to have more frequent visits due to past outcomes, administered treatments, or worsening symptoms (provided that these are recorded in $X$).

    We also consider a stricter variant, which we will refer to as the \textit{strong SAR assumption}: here we assume that $        \lambda(t)dt=\mathbb{P}(dN(t)\!=\!1|  \bar{X}(T),  \bar{N}(T), \bar{A}(T),  \bar{Y}(T)) = \mathbb{P}(dN(t)\!=\!1| \bar{X}^{o}(t^-), \bar{N}(t^-),\bar{A}^{o}(t^-),  \bar{Y}^{o}(t^-))$. This differs from the more general (weaker) SAR assumption above in that observing a patient at time $t$, i.e. $dN(t)$, cannot depend on the covariates $X^o(t)$ to be observed \textit{at} time $t$. As we discuss in the next sections, the weaker SAR assumption already allows identification of treatment effects in our setting, while the strong SAR assumption can greatly simplify estimation of intensities. 
    
    \item 
    \textbf{Sampling not at random (SNAR).} The most general scenario is one where observing is \textit{not} independent of future outcomes conditional on the observed history -- i.e. $\lambda(t)dt\!\! \neq \!\!\mathbb{P}(dN(t)\!=\!1|\bar{X}^{o}(t), \bar{N}(t^-),\bar{A}^{o}(t^-),  \bar{Y}^{o}(t^-))$. This would be the case, e.g., if patients chose to visit due to worsening symptoms that are \textit{not recorded} in $X$ and hence act as a latent cause of the intensity \textit{and} outcome. In this scenario, outcomes cannot be consistently forecast unless further assumptions regarding the sampling or outcome-generating mechanism are made. Therefore, we rely on sampling at random in this work.
\end{itemize}

\section{Forecasting Treatment Outcomes Under Informative Sampling: Goals, Assumptions and Inherent Challenges}\label{sec:capo}

\subsection{Goal: Forecasting treatment outcomes} We aim to estimate \textbf{conditional average potential outcomes (CAPOs)} $\mu_{a,t}(\tau)$ at a future time $t + \tau, \tau \in (0, \tau_\text{max}]$ (with $\tau_\text{max}\leq T-t$): 
 \begin{equation}
  \mu_{a,t}(\tau) =   \mathbb{E}[Y_{{a}}(t\!+\!\tau)|\bar{X}^{o}(t), \bar{N}(t), \bar{A}^{o}(t),  \bar{Y}^{o}(t)]
 \end{equation}
i.e., the instance's expected outcome under treatment plan $a$ conditional on its full observed history $\mathcal{H}^o(t)=\{\bar{X}^{o}(t), \bar{N}(t), \bar{A}^{o}(t),  \bar{Y}^{o}(t)\}$ up to time $t$. We only consider viable treatment plans $a$ subject to $a(t^{*})=A(t^{*})$ for $t^{*}\leq t$ -- i.e., those that do not modify the past, factual treatment history prior to the current time $t$. Such an estimate could be used in practice to decide between competing treatment plans based on expected outcome under either choice. In line with \citet{gische2021forecasting}, we purposefully use the term \textit{forecasting} instead of \textit{predicting} throughout to signify that we wish to give \textit{causal interpretation} to the modeled effects of treatments. This is because, analogously to the standard static setting, unless we make further identifying assumptions, we can in general not assume that predictions based on expectations of the form $\mathbb{E}[Y^o(t+\tau)| A=a, \mathcal{H}^o(t) ]$ are equal to forecasts based on expectations of the form $\mathbb{E}[Y_a(t+\tau)| \mathcal{H}^o(t)]$.

\subsection{Identifying assumptions}
To ensure \textit{identification} of causal claims from observational data, we need to introduce additional assumptions. First, we make assumptions that correspond to adaptations of the  standard \textit{ignorability} assumptions \cite{rubin2005causal} from the standard static setting to our setting. To do so, we define \textit{treatment propensities} for single time-steps $\pi(a(t)) = \mathbb{P}(A(t)\!=\! a(t)|\bar{X}(T), \bar{Y}(T), \bar{A}(T), \bar{N}(T))$ and entire trajectories $\pi_t(a) = \mathbb{P}(A\!=\! a|\mathcal{H}^o(t))$ given history until time $t$.
\begin{assumption}{\textbf{Consistency.}}
Given an observed treatment path $A$, we observe the outcome corresponding to the associated potential outcome:
    $Y^{o}(t) = Y_A(\max s: 0 \leq s \leq t, dN(t)=1)$.
\end{assumption}
\begin{assumption}{\textbf{Unconfoundedness.}} The treatment propensity $\pi(a, t)$ does not depend on future outcomes or unobserved information: 
\begin{equation*}
\begin{split}
\pi(a(t)) = \mathbb{P}(A(t)\!=\! a(t)|\bar{X}(T), \bar{Y}(T), \bar{A}(T), \bar{N}(T)) \\
 = \mathbb{P}(A(t)\!=\!a(t)| \bar{X}^{o}(t),\bar{N}(t), \bar{A}^{o}(t^-), \bar{Y}^{o}(t^-))
\end{split}
\end{equation*}
\end{assumption}
\begin{assumption}{\textbf{Overlap (Positivity for treatment).}} 
    $0 < \mathbb{P}(A = a | \mathcal{H}^o(t)) < 1$, for all admissible treatment paths $a$  and histories $\mathcal{H}^o(t)$ of interest for forecasting.
\end{assumption}
These assumptions are required regardless of the sampling mechanism. For regular sampling, these reduce to the sequential ignorability assumptions made in earlier work \citep[e.g., ][]{lim2018rmsn, bica2019estimating,melnychuk2022causaltransf}.

On top of these ignorability assumptions, estimating causal effects under informative sampling requires making additional assumptions regarding the sampling mechansism \citep{robins1995analysis}. In contrast to existing work which \textit{implicitly} assumed regular observations \citep[e.g., ][]{lim2018rmsn, bica2019estimating} or sampling completely at random \citep[e.g., ][]{seedat2022continuous}, we \textit{explicitly} state our assumed observation process. Specifically, we rely on the weaker, previously introduced \textit{sampling at random} (SAR) assumption:
\begin{assumption}{\textbf{Sampling at random (SAR).}}
    The sampling intensity process does not depend on unobserved or future information, i.e., 
    $\lambda(t)= \mathbb{P}(dN(t)=1| \bar{X}^{o}(t), \bar{N}(t^-),\bar{A}^{o}(t^-),  \bar{Y}^{o}(t^-))$.
\end{assumption}

Analogous to assumptions on treatment overlap, we assume the probability of observing at any point in time is bounded away from zero, for any history of interest for forecasting:
\begin{assumption}{\textbf{Positivity of observation.}}
$\mathbb{P}(dN(t+\tau)=1|  \mathcal{H}^o(t) ) > 0$ for any $ \tau \in (0, T-t]$ and history $\mathcal{H}^o(t)$ of interest.
\end{assumption}

Finally, we assume that all treatment events are observed. In most applications, this is a natural assumption (e.g., if treatments are administered at a hospital). It is generally not possible to estimate treatment effects from observed outcomes without knowing which treatments were administered, unless further assumptions are made \citep{kennedy2020efficient}.
\begin{assumption}{\textbf{Observed treatments.}} All treatments are observed, i.e. $\mathbb{P}(A(t)=1|dN(t)=0)=0$.
\end{assumption}

The identifying assumptions discussed above can equivalently be expressed as a generative model, determining the temporal ordering of realizations of the different observed variables\footnote{In principle, other generative models could be assumed as long as sufficient exclusion restrictions between observation-/treatment-generating processes and outcome-generating processes are made. For example, the visit process can depend on future treatments if such treatments are pre-scheduled.}. In particular, at each time $t$, the visit decision $dN(t)$ is realized first, which can depend on observed histories $\mathcal{H}^o(t^-)$ and covariates to be observed $X^o(t)$ (SAR) or on  $\mathcal{H}^o(t^-)$ only (strong SAR); the former implies a setting where e.g. patients present themselves for an appointment due to worsening symptoms while the latter allows only scheduling of future appointments due to symptoms already observed earlier. If $dN(t)=1$, then covariates $X^o(t)$ are first observed, treatment $A(t)$ is then determined based only on observed information ($\mathcal{H}^o(t^-)$, $N(t)$ \&  $X^o(t)$) and, finally, the outcome is realized and observed as  $Y^o(t)$.

\subsection{What makes learning CAPOs from observational data challenging?}\label{sec:covshift}
If we had access to the complete data structure with all potential outcomes $Y_a(t)$, learning an estimate  $\hat{\mu}_{a, t}(\tau; \mathcal{H}^o(t))$ for the CAPOs with fixed $a$ would be a standard ML problem: we would search a hypothesis function in some hypothesis class $\mathcal{F}$ that minimizes the expected (oracle) risk $R^{*}$, i.e.,
\begin{equation}
    \hat{\mu}_{a, t}(\tau; \mathcal{H}^o(t)) \in \arg \min_{f_{a,\tau} \in \mathcal{F}}R^{*}(f_{a,\tau})
\end{equation}
where, for some loss function $\ell$, and using the shorthands $t'= t + \tau$ and $h_t = \mathcal{H}^o(t)$
\begin{equation*}
\begin{split}
R^{*}(f_{a,\tau})=  \mathbb{E}\left[\int^T_0 \!\! \int^{\tau_\text{max}}_t \!\!\! \! \! \ell\left(Y_a(t'), f_{a,\tau}(h_t)\right)d\tau dt\right] =
\\ \int^T_0 \!\! \int^{\tau_\text{max}}_0 \!\!\! \! \!\int \! \! \!\int \ell\left(y_a(t'), f_{a,\tau}(h_t)\right) d{P}(y_a(t')|h_t)d{P}(h_t)d\tau dt
\end{split}
\end{equation*}

However, as previously discussed, in reality we only have access to observational data in which patients are (i) incompletely, irregularly, and informatively observed and (ii) characterized by only a single \textit{factual} outcome corresponding to the treatment actually received. If we were to learn a standard ML predictor from this observed data, we would instead be optimizing the observational risk  $R^{obs}(h_{a,\tau})=$
\begin{equation*}
\begin{split}
\mathbb{E}\left[\int^T_0 \!\! \int^{\tau_\text{max}}_0 \!\!\! \! \! \mathbbm{1}\{A\!=\!a\}dN(t')\ell\left(Y_a(t'), f_{a,\tau}(h_t)\right)d\tau dt\right] = \\ \int^T_0 \!\! \int^{\tau_\text{max}}_t \!\!\! \! \!\int \! \! \!\int \ell(y_a(t'), f_{a,\tau}(h_t)) \textcolor{green!80!black}{\pi_t(a)} \textcolor{blue!50}{\lambda_t(t')} \\ d{P}(y_a(t')|h_t)d{P}(h_t)d\tau dt
\end{split}
\end{equation*}

Thus, unless the $\tau$-step ahead intensity $\textcolor{blue!50}{\lambda_t(t')}$, defined through $\lambda_t(t')d\tau = \mathbb{P}(dN(t\!+\!\tau)\!=\!1|\mathcal{H}^o(t) \cup \bar{A}(t+\tau^-))$, and treatment propensity $\textcolor{green!80!black}{\pi_t(a)}$ are constant across patient histories, the minimizers of $R^*$ and $R^{obs}$ will in general be different. Intuitively, this is because the distribution of patient characteristics in the observed data can differ from the distribution in the underlying unobserved complete distribution, both due to informative sampling and treatment selection. Thus, the challenge we are facing here is one of \textit{covariate shift} between the training data and hypothetical test data. 
{\parfillskip=0pt \emergencystretch=.5\textwidth \par}

Covariate shift and its potential remedies have been studied in much depth in the recent ML literature (see e.g. \citet{redko2020survey, farahani2021brief} for recent overviews). Here, we explore the use of one of the oldest and most well-established solutions: \textit{importance weighting} \cite{shimodaira2000improving}. That is, as further discussed in the next section, we propose to minimize a weighted observational risk 
\begin{equation}
R^{w}(f_{a,\tau})=\mathbb{E}\left[\int^T_0 \!\! \int^{\tau_\text{max}}_0 w_{a,\tau} \ell(f_{a, \tau})  d\tau dt \right]
\end{equation}
with $\textstyle \ell(f_{a, \tau})=\mathbbm{1}\{A\!=\!a\}dN(t')\ell\left(Y_a(t'), f_{a,\tau}(h_t)\right)$. For oracle importance weights $w^*_{a,\tau}= \frac{1}{\textcolor{green!80!black}{\pi(a)} \textcolor{blue!50}{\lambda_t(t')}}$ it is easy to see that $R^{obs, w^*}(f_{a,\tau}) = R^{*}(f_{a,\tau})$.

\section{Learning to Forecast Treatment Outcomes Under Informative Sampling}
This section presents a methodology for learning to forecast treatment outcomes under informative sampling. \cref{ssec:general_framework} presents a general framework that is compatible with any ML algorithm capable of predicting outcomes over time. \cref{ssec:method_implementation} instantiates this framework using Neural CDEs.
{\parfillskip=0pt \emergencystretch=.5\textwidth \par}

\subsection{Learning To Forecast Using Inverse Intensity Weights}\label{ssec:general_framework}

The analysis presented in \cref{sec:covshift} allows straightforward construction of a framework for learning to forecast treatment outcomes from informatively sampled (SAR) data. Given an ML algorithm $\mathcal{A}$ that can issue continuous-time predictions using irregularly sampled data, one simply needs to fit $\mathcal{A}$ on the observed data while providing appropriate importance weights $w$. As the true weights will generally be unknown in practice, one might have to use $\mathcal{A}$ to also learn (i) observation intensities and (ii) treatment propensities to gain access to estimates of the true importance weights. As we discuss for a specific example in \cref{sec:weight}, one could learn such weights either in a pre-processing step or in an end-to-end fashion.

When learning intensity weights, it becomes important whether one makes the general SAR or the strong SAR assumption: under the strong SAR assumption, learning $\lambda_t(t')$ comes down to the easier task of estimating $\mathbb{P}(dN(t')=1|\mathcal{H}^o(t))$ directly, where $t'=t+\tau$. Under the more general SAR assumption, one needs to model $dN(t')$ and $X^o(t')$ \textit{jointly} as a \textit{marked} point process to learn the distributions ${P}(dN(t'), X^o(t')|\mathcal{H}^o(t))={P}(dN(t')|X^o(t'), \mathcal{H}^o(t)){P}(X^o(t')|\mathcal{H}^o(t))$, where ${P}(X^o(t')|\mathcal{H}^o(t))$ could be a high-dimensional continuous density. In the remainder, we therefore restrict ourselves to the strong SAR setting -- allowing us to highlight the challenges arising when learning under some form of informative sampling. It would be an interesting next step to incorporate some of the recent work on Neural Temporal Point Processes \citep[see e.g. ][]{shchur2021neural} to enable learning under more complex dependencies.  
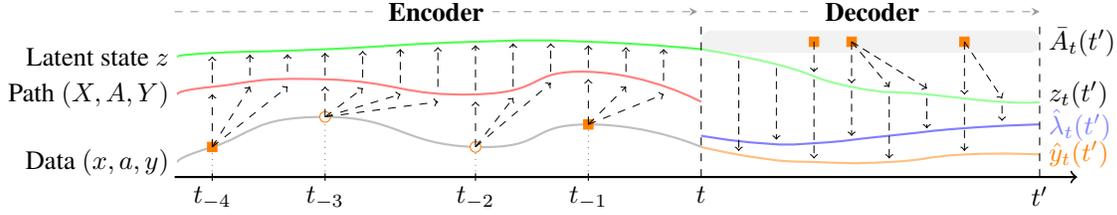
\begin{figure*}[!t]
\centering
\begin{tikzpicture}[square/.style={regular polygon, regular polygon sides=4}]
    \draw [thick, ->] (0,0) -- (12,0) node [mid right] {};
    \coordinate (start_x) at (0.05,0.2);
    \coordinate (x1) at (0.5,0.4);
    \coordinate (x2) at (2,0.8);
    \coordinate (x3) at (4,0.4);
    \coordinate (x4) at (5.5,0.7);
    \coordinate (end_x) at (7,0.4);
    
    \coordinate (start_path) at (0.05,1.1);
    \coordinate (p1) at (0.5,1.2);
    \coordinate (p2) at (2,1.3);
    \coordinate (p3) at (4,1.1);
    \coordinate (p4) at (5.5,1.4);
    \coordinate (end_path) at (7,1.0);
    
    \coordinate (start_z) at (0.05,1.6);
    \coordinate (z1) at (0.5,1.65);
    \coordinate (z2) at (2,1.7);
    \coordinate (z3) at (4,1.8);
    \coordinate (z4) at (5.5,1.8);
    \coordinate (end_z) at (7,1.7);
    
    \coordinate (start_y) at (7, 0.4);
    \coordinate (y1) at (7.8,0.25);
    \coordinate (y2) at (8.5,0.2);
    \coordinate (y3) at (9.5,0.2);
    \coordinate (y4) at (10.5,0.3);
    \coordinate (end_y) at (11.5,0.3);
    
    \coordinate (start_l) at (7, 0.55);
    \coordinate (l1) at (7.8,0.45);
    \coordinate (l2) at (8.5,0.45);
    \coordinate (l3) at (9.5,0.55);
    \coordinate (l4) at (10.5,0.65);
    \coordinate (end_l) at (11.5,0.7);
    
    \coordinate (start_s) at (7,1.7);
    \coordinate (s1) at (8,1.5);
    \coordinate (s2) at (9,1.2);
    \coordinate (s3) at (10,1.1);
    \coordinate (s4) at (11,1.0);
    \coordinate (end_s) at (11.5,1.0);

    \coordinate (treat1) at (8.5, 1.8);
    \coordinate (treat2) at (9, 1.8);
    \coordinate (treat3) at (10.5, 1.8);
    \draw[rounded corners, fill=gray!10, draw=none] (7, 1.65) rectangle (11.5, 1.95) {};
    \foreach \n in {treat1, treat2, treat3} \draw node[square, draw=orange, fill=orange, inner sep=0pt, minimum width=5pt, minimum height=5pt] (d) at (\n) {};
    \draw [thick, gray!50] plot[smooth, tension=1]
    coordinates{(start_x) (x1) (x2) (x3) (x4) (end_x)};
    \draw [thick, red!50] plot[smooth, tension=1]
    coordinates{(start_path) (p1) (p2) (p3) (p4) (end_path)};
    \draw [thick, green!70] plot[smooth, tension=1]
    coordinates{(start_z) (z1) (z2) (z3) (z4) (end_z)};
    \draw [thick, green!40] plot[smooth, tension=1]
    coordinates{(start_s) (s1) (s2) (s3) (s4) (end_s)};
    \draw [thick, orange!50] plot[smooth, tension=1]
    coordinates{(start_y) (y1) (y2) (y3) (y4) (end_y)};
    \draw [thick, blue!50] plot[smooth, tension=1]
    coordinates{(start_l) (l1) (l2) (l3) (l4) (end_l)};
    \foreach \c in {x1,x2,x3,x4} {
        \draw [very thin, dotted] let \p1=(\c) in (\x1,0) -- (\c);
    };
    \foreach \c in {x1,x2,x3,x4} {
        \draw [thin, solid] let \p1=(\c) in (\x1,-0.05) -- (\x1,0.05);
    };
    \draw [thin, dashed] (7, 0) -- (7, 2);
    \draw [thin, dashed] (11.5, 0) -- (11.5, 2);
    \draw let \p1 = (x1) in node at (\x1, 0) [below] {$t_{-4}$};
    \draw let \p1 = (x2) in node at (\x1, 0) [below] {$t_{-3}$};
    \draw let \p1 = (x3) in node at (\x1, 0) [below] {$t_{-2}$};
    \draw let \p1 = (x4) in node at (\x1, 0) [below] {$t_{-1}$};
    \draw node at (7, 0) [below] {$t$};
    \draw node at (11.5, 0) [below] {$t'$};
    \foreach \n in {x1,x4} \draw node[square, draw=orange, fill=orange, inner sep=0pt, minimum width=5pt, minimum height=5pt] (d) at (\n) {};
    \foreach \n in {x2,x3} \fill [draw=orange, fill=white] (\n) circle (2pt) node [above] {};
    \draw [->, very thin, dashed] ([shift=({0,0})]x1) -- ([shift=({0,-0.1})]p1);
    \draw [->, very thin, dashed] ([shift=({0,0})]x1) -- ([shift=({0.5,0})]p1);
    \draw [->, very thin, dashed] ([shift=({0,0})]x1) -- ([shift=({1,0})]p1);
    \draw [->, very thin, dashed] ([shift=({0,0})]x2) -- ([shift=({0,-0.1})]p2);
    \draw [->, very thin, dashed] ([shift=({0,0})]x2) -- ([shift=({0.5,-0.12})]p2);
    \draw [->, very thin, dashed] ([shift=({0,0})]x2) -- ([shift=({1,-0.20})]p2);
    \draw [->, very thin, dashed] ([shift=({0,0})]x2) -- ([shift=({1.5,-0.3})]p2);
    \draw [->, very thin, dashed] ([shift=({0,0})]x3) -- ([shift=({0,-0.1})]p3);
    \draw [->, very thin, dashed] ([shift=({0,0})]x3) -- ([shift=({0.5,-0.05})]p3);
    \draw [->, very thin, dashed] ([shift=({0,0})]x3) -- ([shift=({1,0.15})]p3);
    \draw [->, very thin, dashed] ([shift=({0,0})]x4) -- ([shift=({0,-0.1})]p4);
    \draw [->, very thin, dashed] ([shift=({0,0})]x4) -- ([shift=({0.5,-0.2})]p4);
    \draw [->, very thin, dashed] ([shift=({0,0})]x4) -- ([shift=({1,-0.3})]p4);
    \draw [->, very thin, densely dashed] ([shift=({0,0.05})]p1) -- ([shift=({0,-0.05})]z1);
    \draw [->, very thin, densely dashed] ([shift=({0.5,0.15})]p1) -- ([shift=({0.5,-0.025})]z1);
    \draw [->, very thin, densely dashed] ([shift=({1,0.2})]p1) -- ([shift=({1,-0.02})]z1);
    \draw [->, very thin, densely dashed] ([shift=({0,0.05})]p2) -- ([shift=({0,-0.05})]z2);
    \draw [->, very thin, densely dashed] ([shift=({0.5,0.025})]p2) -- ([shift=({0.5,-0.025})]z2);
    \draw [->, very thin, densely dashed] ([shift=({1,-0.05})]p2) -- ([shift=({1,0.0})]z2);
    \draw [->, very thin, densely dashed] ([shift=({1.5,-0.1})]p2) -- ([shift=({1.5,0.025})]z2);
    \draw [->, very thin, densely dashed] ([shift=({0,0.05})]p3) -- ([shift=({0,-0.05})]z3);
    \draw [->, very thin, densely dashed] ([shift=({0.5,0.1})]p3) -- ([shift=({0.5,-0.025})]z3);
    \draw [->, very thin, densely dashed] ([shift=({1,0.3})]p3) -- ([shift=({1,-0.025})]z3);
    \draw [->, very thin, densely dashed] ([shift=({0,0.05})]p4) -- ([shift=({0,-0.05})]z4);
    \draw [->, very thin, densely dashed] ([shift=({0.5,-0.0})]p4) -- ([shift=({0.5,-0.075})]z4);
    \draw [->, very thin, densely dashed] ([shift=({1,-0.12})]p4) -- ([shift=({1,-0.1})]z4);
    \draw [->, very thin, densely dashed] ([shift=({0,0})]treat1) -- ([shift=({0,-0.4})]treat1);
    \draw [->, very thin, densely dashed] ([shift=({0,0})]treat2) -- ([shift=({0,-0.55})]treat2);
    \draw [->, very thin, densely dashed] ([shift=({0,0})]treat2) -- ([shift=({0.5,-0.6})]treat2);
    \draw [->, very thin, densely dashed] ([shift=({0,0})]treat2) -- ([shift=({1,-0.62})]treat2);
    \draw [->, very thin, densely dashed] ([shift=({0,0})]treat3) -- ([shift=({0,-0.7})]treat3);
    \draw [->, very thin, densely dashed] ([shift=({0,0})]treat3) -- ([shift=({0.5,-0.75})]treat3);
    \draw [->, very thin, densely dashed] ([shift=({-0.5,0.05})]s1) -- ([shift=({-0.5,-1.15})]s1);
    \draw [->, very thin, densely dashed] ([shift=({0,-0.05})]s1) -- ([shift=({0,-1.0})]s1);
    \draw [->, very thin, densely dashed] ([shift=({-0.5,0.1})]s2) -- ([shift=({-0.5,-0.95})]s2);
    \draw [->, very thin, densely dashed] ([shift=({0,-0.05})]s2) -- ([shift=({0,-0.65})]s2);
    \draw [->, very thin, densely dashed] ([shift=({-0.5,-0.025})]s3) -- ([shift=({-0.5,-0.85})]s3);
    \draw [->, very thin, densely dashed] ([shift=({0,-0.05})]s3) -- ([shift=({0,-0.45})]s3);
    \draw [->, very thin, densely dashed] ([shift=({-0.5,-0.025})]s4) -- ([shift=({-0.5,-0.65})]s4);
    \draw [->, very thin, densely dashed] ([shift=({0,-0.05})]s4) -- ([shift=({0,-0.25})]s4);
    \fill [draw=none, fill=none, text=black] (start_x) circle (2pt) node [left] {Data $(x, a, y)$};
    \fill [draw=none, fill=none, text=black] (start_path) circle (2pt) node [left] {Path $(X, A, Y)$};
    \fill [draw=none, fill=none, text=black] (start_z) circle (2pt) node [left] {Latent state $z$};
    \fill [draw=none, fill=none, text=black] (11.5, 1.8) circle (2pt) node [right] {$\Bar{A}_t(t')$};
    \fill [draw=none, fill=none, text=black] ([shift=({0,0.1})]end_s) circle (2pt) node [right] {$z_t(t')$};
    \fill [draw=none, fill=none, text=blue!50] (end_l) circle (2pt) node [right] {$\hat{\lambda}_t(t')$};
    \fill [draw=none, fill=none, text=orange] (end_y) circle (2pt) node [right] {$\hat{y}_t(t')$};

    \draw [thin, -stealth, dashed, opacity=0.4] (0,2.2) -- (6.95,2.2) node [midway,fill=white, opacity=1] {\textbf{Encoder}};
    \draw [thin, -stealth, dashed, opacity=0.4] (7.05,2.2) -- (11.5,2.2) node [midway,fill=white, opacity=1] {\textbf{Decoder}};
\end{tikzpicture}
\caption{\textbf{TESAR-CDE: Adapting TE-CDE for learning given SAR.} The history of observations (\observation) and treatments (\treatment) up to time $t$ is first encoded as a continuous latent path $z(t)$. Based on a future treatment plan $\Bar{A}_t(t')$, the decoder then forecasts a future latent path $z_t(t')$, with $t' = t + \tau$. In contrast to TE-CDE, TESAR-CDE (1) uses the latent path $z_t(t')$ to forecast both the outcome $\textcolor{orange}{\hat{y}_t(t')}$ and intensity $\textcolor{blue!50}{\hat{\lambda}_t(t')}$, and (2) uses the intensity to weight the outcome loss using $\mathcal{L}^\text{WMSE}$.}
\vspace{-2pt}
\hrulefill
\vspace{-12pt}
\label{fig:neural_CDE}
\end{figure*}

In the following, we discuss two possible implementations of this framework using Neural CDEs by extending the methodology for learning continuous time treatment effects of \citet{seedat2022continuous} (which originally did not correct for informative sampling). Nevertheless, the approach discussed above is more general and could be applied to any ML model that can predict $Y(t)$.

\subsection{TESAR-CDE: Forecasting with Intensity-weighted Neural CDEs}\label{ssec:method_implementation}
This section presents TESAR-CDE, a specific implementation of the framework discussed above by extending TE-CDE \citep{seedat2022continuous} to account for informative sampling (SAR), see \cref{fig:neural_CDE} for a graphical overview. In the remainder of this work, we focus on the special case where complete treatment plans $A$ are \textit{randomly} assigned and \textit{fixed} at time $t=0$; such a situation commonly arises in practice, e.g., in a clinical trial with a dynamic observation plan \citep{lin2004analysis, buuvzkova2009semiparametric}. This allows to \textit{single out} the challenges arising solely due to the presence of informative observations. Moreover, this allows us to highlight that the forecasting problem remains challenging even in the absence of all treatment selection bias (the main challenge addressed in related work). Nevertheless, if required, any existing method equipped to deal with outcome-treatment confounding -- e.g., using importance weighting \cite{lim2018rmsn} or adversarial training \cite{seedat2022continuous} -- could simply be combined with the inverse intensity weighting approaches we discuss and test below.

\subsubsection{Background: TE-CDE} Treatment Effect Neural Controlled Differential Equation \citep[TE-CDE;][]{seedat2022continuous} is a recently proposed model for forecasting treatment effects from irregularly sampled data. TE-CDE views observations as samples from an underlying continuous-time process and uses Neural CDEs \citep{kidger2020neural} to learn this latent trajectory. First, an encoder learns a latent path $z(t)$ as the solution of a CDE:
\begin{equation}
    z(t_0) = g(X(t_0), A(t_0), Y(t_0)),
\notag
\end{equation}
\begin{equation}
    z(t) = z(t_0) + \int_{t_0}^t f_\theta(z(s)) \frac{d(X(s), A(s), Y(s))}{ds} ds
\notag
\end{equation}
with $g$ and $f_\theta$ neural networks. This is achieved by solving the above initial value problem (IVP), $\forall s \in [t_0, t]$:
\begin{equation}
    z(t) = \texttt{ODESolve}(f_\theta, z(t_0), \Bar{X}(t), \Bar{A}(t), \Bar{Y}(t))
\notag
\end{equation}
using a numerical ODE solver \citep{kidger2020neural}. The decoder forecasts the future latent path $z_t(t')$ by solving a second IVP given the future treatment plan $\Bar{A}_t(t')$:
\begin{equation}    
    z_t(t') = \texttt{ODESolve}(f_\phi, z(t), \Bar{A}_t(t'))
    ,
\notag
\end{equation}
with decoder network $f_\phi$ and $t' = t + \tau$. A final network $f_\psi$ maps the latent path $z_t(t')$ to the outcome $\hat{y}_t(t') = f_\psi(z_t(t'))$. \cref{sfig:TE_CDE} shows the complete architecture.

The entire model ($g, f_\theta$, $f_\phi$ and $f_\psi$) is trained by optimizing the mean squared error (MSE) of the predicted outcome:
\begin{equation}
\begin{split}
    \mathcal{L}^{\text{MSE}}_i \! = \!\! \int_{0}^T \!\!\!\! \int_{0}^{\tau_\text{max} } \!\!\!\!\! dN_i(t') \left(y_{i}(t') - \hat{y}_{i, t}(t') \right)^2 dt d\tau 
    .
\notag
\end{split}
\end{equation}
This way, the mean squared error is calculated using the observed outcomes in the considered forecasting horizon $(0, \tau_\text{max}]$, for each timestep $t \in [0, T]$.
To account for bias resulting from time-dependent confounding, TE-CDE also uses domain adversarial training to learn a treatment-invariant representation. However, as discussed above, we focus on unconfounded settings in the remainder of this work and therefore do not include this, though it is straightforward to add it in settings where needed.

\subsubsection{TESAR-CDE: Learning to forecast with an inverse-intensity weighted loss}\label{sec:weight} 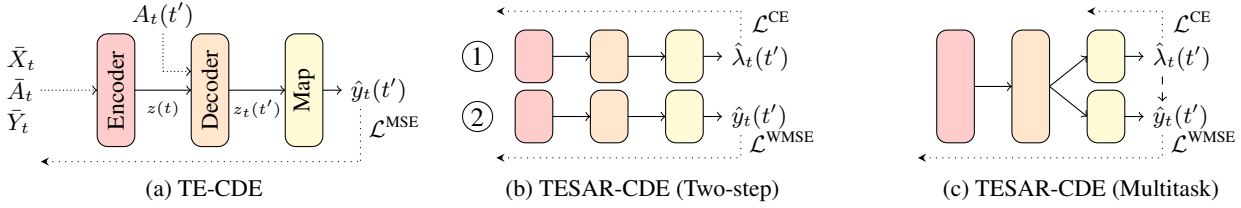
\begin{figure*}[!t]
\centering
\begin{subfigure}[b]{0.32\textwidth}
    \centering
    \begin{tikzpicture}
        \node at (-0.2, 0.75) [text width=0.5cm] {\small $\Bar{X}_t$ $\Bar{A}_t$ $\Bar{Y}_t$};
        \draw[->, densely dotted] (0, 0.75) -- (0.75, 0.75);
        \draw[draw=black, rounded corners, fill=red!20] (0.75, 0) rectangle ++(0.5, 1.5);
        \filldraw (1, 0.75) node [rotate=90] {\small Encoder};
        \node at (1.625, 0.5) {\tiny $z(t)$};
        \node at (1.625, 1.75) {\small $A_t(t')$};
        \draw[-, densely dotted] (1.625, 1.5) -- (1.625, 1);
        \draw[->, densely dotted] (1.625, 1) -- (2, 1);
        \draw[->] (1.25, 0.75) -- (2.0, 0.75);
        \draw[draw=black, rounded corners, fill=orange!20] (2.0, 0) rectangle ++(0.5, 1.5);
        \filldraw (2.25, 0.75) node [rotate=90] {\small Decoder};
        \node at (2.875, 0.5) {\tiny $z_t(t')$};
        \draw[->] (2.5, 0.75) -- (3.25, 0.75);
        \draw[draw=black, rounded corners, fill=yellow!20] (3.25, 0) rectangle ++(0.5, 1.5);
        \filldraw (3.5, 0.75) node [rotate=90] {\small Map};
        \draw[->] (3.75, 0.75) -- (4, 0.75);
        \filldraw (4, 0.75) node [right] {\small $\hat{y}_t(t')$};
        
        \draw[stealth-, dotted] (0, -0.2) -- (4.25, -0.2);
        \draw[-, dotted] (4.25, 0.5) -- (4.25, -0.2);
        \filldraw (4.7, 0) node [above] {\small $\mathcal{L}^{\text{MSE}}$};
    \end{tikzpicture}
    \caption{{TE-CDE}}
    \label{sfig:TE_CDE}
\end{subfigure}
\hfill
\begin{subfigure}[b]{0.32\textwidth}
    \centering
    \begin{tikzpicture}
        \draw (0.25, 1.15) circle [radius=0.2] node {1};
        \draw (0.25, 0.35) circle [radius=0.2] node {2};
        
        \draw[draw=black, rounded corners, fill=red!20] (0.75, 0.8) rectangle ++(0.5, 0.7);
        \draw[draw=black, rounded corners, fill=red!20] (0.75, 0) rectangle ++(0.5, 0.7);
        \draw[->] (1.25, 1.15) -- (1.75, 1.15);
        \draw[->] (1.25, 0.35) -- (1.75, 0.35);
        \draw[draw=black, rounded corners, fill=orange!20] (1.75, 0.8) rectangle ++(0.5, 0.7);
        \draw[draw=black, rounded corners, fill=orange!20] (1.75, 0) rectangle ++(0.5, 0.7);
        \draw[->] (2.25, 1.15) -- (2.75, 1.15);
        \draw[->] (2.25, 0.35) -- (2.75, 0.35);
        \draw[draw=black, rounded corners, fill=yellow!20] (2.75, 0.8) rectangle ++(0.5, 0.7);
        \draw[draw=black, rounded corners, fill=yellow!20] (2.75, 0) rectangle ++(0.5, 0.7);
        \draw[->] (3.25, 1.15) -- (3.5, 1.15);
        \filldraw (3.5, 1.15) node [right] {\small $\hat{\lambda}_t(t')$};
        \draw[->] (3.25, 0.35) -- (3.5, 0.35);
        \filldraw (3.5, 0.35) node [right] {\small $\hat{y}_t(t')$};

        \draw[stealth-, dotted] (0.5, -0.2) -- (3.75, -0.2);
        \draw[-, dotted] (3.75, 0.1) -- (3.75, -0.2);
        \filldraw (4.15, 1.35) node [above] {\small $\mathcal{L}^{\text{CE}}$};
        \filldraw (4.3, -0.25) node [above] {\small $\mathcal{L}^{\text{WMSE}}$};
        \draw[stealth-, dotted] (0.5, 1.75) -- (3.75, 1.75);
        \draw[-, dotted] (3.75, 1.4) -- (3.75, 1.75);
    \end{tikzpicture}
    \caption{{TESAR-CDE (Two-step)}}
    \label{sfig:Two_step}
\end{subfigure}
\hfill
\begin{subfigure}[b]{0.32\textwidth}
    \centering
    \begin{tikzpicture}
        \draw[draw=black, rounded corners, fill=red!20] (0.75, 0) rectangle ++(0.5, 1.5);
        \draw[->] (1.25, 0.75) -- (1.75, 0.75);
        \draw[draw=black, rounded corners, fill=orange!20] (1.75, 0) rectangle ++(0.5, 1.5);
        \draw[->] (2.25, 0.75) -- (2.75, 1.15);
        \draw[->] (2.25, 0.75) -- (2.75, 0.35);
        \draw[draw=black, rounded corners, fill=yellow!20] (2.75, 0.8) rectangle ++(0.5, 0.7);
        \draw[draw=black, rounded corners, fill=yellow!20] (2.75, 0) rectangle ++(0.5, 0.7);
        \draw[->, out=0, in=90] (3.25, 1.15) -- (3.5, 1.15);
        
        \draw[->] (3.25, 0.35) -- (3.5, 0.35);
        \filldraw (3.5, 1.15) node [right] {\small $\hat{\lambda}_t(t')$};
        \filldraw (3.5, 0.35) node [right] {\small $\hat{y}_t(t')$};

        \draw[<-, dashed] (3.75, 0.55) -- (3.75, 0.95);
        \draw[stealth-, dotted] (2.75, 1.75) -- (3.75, 1.75);
        \draw[-, dotted] (3.75, 1.4) -- (3.75, 1.75);
        \filldraw (4.15, 1.35) node [above] {\small $\mathcal{L}^{\text{CE}}$};
        \draw[stealth-, dotted] (0.5, -0.2) -- (3.75, -0.2);
        \draw[-, dotted] (3.75, 0.1) -- (3.75, -0.2);
        \filldraw (4.3, -0.25) node [above] {\small $\mathcal{L}^{\text{WMSE}}$};
        
    \end{tikzpicture}
    \caption{{TESAR-CDE (Multitask)}}
    \label{sfig:Multitask}
\end{subfigure}
\caption{\textbf{Comparing TESAR-CDE to TE-CDE.} We show TE-CDE and our proposed alternative, TESAR-CDE, in its two-step and multitask configuration. Arrows indicate the input (\inputarrow), forward pass (\forwardarrow) and backpropagation (\backproparrow). The multitask model uses the intensity loss only to train the intensity map, but not the shared encoder or decoder. The dashed arrow (\nobackproparrow) indicates that the intensities are used as weights $\lambda_t^{-1}$ in $\mathcal{L}^{\text{WMSE}}$, but not backpropagated as part of this loss.}
\vspace{-2pt}
\hrulefill
\vspace{-12pt}
\label{fig:architectures}
\end{figure*}

In this section, we instantiate our previously proposed framework using Neural CDEs, resulting in TESAR-CDE, Treatment Effects given Sampling At Random using Neural CDEs. Essentially, we extend TE-CDE for learning under informative sampling. Given (estimated) intensities $\hat{\lambda}_{i,t}(t')$, adapting TE-CDE's outcome loss to adjust for informative observations is straightforward: the inverse of these estimated intensities can be used as importance weights to train TE-CDE for outcome prediction using a \textit{weighted} MSE:
\begin{equation}
\begin{split}
    \mathcal{L}^{\text{WMSE}}_i \! = \!\! \int_{0}^T \!\!\!\! \int_{0}^{\tau_\text{max} } \!\!\!\!\! dN_i(t') \, \frac{\left(y_{i}(t') - \hat{y}_{i, t}(t') \right)^2}{\textcolor{blue!50}{\hat{\lambda}_{i, t}(t')}} \, dt d\tau\!
    .
\end{split}
\notag
\end{equation}

We propose a multi-stage or end-to-end version of TESAR-CDE (\cref{fig:architectures} compares the proposed architectures with TE-CDE). Both predict the intensity from $z_t$ as $\hat{\lambda}_{i, t}(t') = f^\lambda_\psi(z_t(t'))$. We assume there is a minimal sampling interval $dt$; e.g., doctors might not measure covariates more than once per hour. Let $dN_{i, t}(t') = 1$ if instance $i$ was observed in interval $(t, t']$, $0$ otherwise. The intensity $\lambda_{i, t}(t')$ can then be approximated by minimizing the cross-entropy
\begin{equation}
\begin{split}
    \mathcal{L}^{\text{CE}}_i = - \! \sum_{t=0}^T \! \sum_{\tau=0}^{\tau_\text{max}} \Big[ & \, dN_{i, t}(t') \log\! \left(\hat{\lambda}_{i, t}(t')\right) \\ & \,\, + \left(1\!-\!dN_{i, t}(t')\right) \log\!\left(1\!-\!\hat{\lambda}_{i, t}(t')\right) \! \Big]
    ,
\end{split}
\notag
\end{equation}
where $t' = t + \tau$.
For applications where no minimal time step $dt$ exists, neural point processes can be used to learn the intensity in continuous time \citep[see e.g.][]{shchur2021neural}.

The \textbf{two-step} procedure consists of two TE-CDE style models that are trained sequentially. A first model predicts the intensity $\lambda_{i, t}(t')$; a second model uses the inverse of these intensities $\lambda_{i, t}(t')^{-1}$ as weights in its weighted MSE loss. Alternatively, we can combine both tasks in a \textbf{multitask} framework to predict both intensities and outcomes:
\begin{equation}
    \mathcal{L}^{\text{MT}} = (1 - \alpha) \mathcal{L}^{\text{WMSE}}_i + \alpha \mathcal{L}^{\text{CE}}_i ,
\end{equation}
with hyperparameter $\alpha$ balancing the importance of the two terms. 
The intensity loss only optimizes the intensity map $f_\psi^{\lambda}$; the weighted MSE is used to optimize the rest of the network ($g, f_\theta, f_\phi, f_\psi^{y}$). Moreover, similar to \citet{hassanpour2019counterfactual}'s architecture for learning importance weights to correct for treatment-outcome confounding, we do not backpropagate with respect to the intensity weights in the weighted MSE for outcome prediction, as this could bias the network to predict small weights (i.e. large intensities) in order to minimize the weighted MSE.

The potential benefits of the multitask framework are threefold. First, learning a shared representation $z_t$ to predict both outcome and intensity results in fewer parameters. Second, it requires only training one network and one call to the ODE solver per iteration, resulting in computational speedups. Third, to reduce the variability due to importance weighting, we only optimize the shared representation of the multitask model for outcome prediction. For bias correction using importance weighting, the shared representation does \textit{not} need to be a sufficient statistic for predicting the intensity. This is because we only need to care about the non-uniformity in observation intensity insofar as it is related to the outcome. The reason for this is conceptually identical to why one should not include predictors of treatment only (a.k.a. instruments) in a propensity score \citep{vanderweele2019principles} and why sufficient dimensionality reduction before importance weighting is recommended in general applications with covariate shift \citep{maia2022effective}: importance weighting generally only needs to adjust for shifts in variables that are themselves predictors of the outcome. Our multitask learner implicitly enforces this by optimizing the shared representation based on the outcome loss only. The two-step and multitask architectures are illustrated in \cref{sfig:Two_step,sfig:Multitask}. \cref{sec:appendix_training,sec:appendix_tesar_cde_implement} provide more details on the training procedure and implementation.


\section{Results}

To assess the impact of informative sampling, we propose a novel simulation environment that allows us to control the level of informativeness and assess its effect on the resulting model's performance. Our simulation is inspired by real-world randomized controlled trials that compared treatment regimes in the context of lung cancer \citep{furuse1999phase, auperin2010meta, curran2011sequential}. Given the patient's history, our goal is to forecast the patient's tumor size for a potential future treatment plan. Our code is available at \href{https://github.com/toonvds/TESAR-CDE}{https://github.com/toonvds/TESAR-CDE}.

\subsection{Simulation: lung cancer treatment}

Following existing work \citep[e.g., ][]{melnychuk2022causaltransf, seedat2022continuous}, we simulate data based on the tumor growth model of \citet{geng2017prediction}. To analyze the effect of informative sampling, we combine this tumor growth model with a sampling mechanism in which the degree of informativeness can be controlled. We refer to \cref{sec:app_simulation} for more detailed information and visualizations.

\textbf{Tumor growth simulation.}
We simulate tumor growth based on a pharmacokinetic-pharmacodynamic model of \citet{geng2017prediction}. This model simulates the outcome, tumor volume $Y(t)$, based on the historical tumor volume, tumor growth, chemotherapy, and radiotherapy:
\begin{align}
    \frac{dY(t)}{dt} = \Big[ 1 
    + \overbrace{\rho \log \left(\frac{K}{Y(t)}\right)}^{\text{Tumor growth}} 
    - & \overbrace{\beta_c C(t)}^{\text{Chemotherapy}}  \\ \notag
    - \underbrace{\left(\alpha_r d(t) + \beta_r d(t)^2 \right)}_{\text{Radiotherapy}}
    &+ \underbrace{\epsilon(t)}_{\text{Noise}} \Big] Y(t),
\end{align}
with $K, \rho, \beta_c, \alpha_r, \beta_r, \epsilon_t$ sampled following \citet{geng2017prediction}; $C(t)$ and $d(t)$ are set following existing work \citep{lim2018rmsn, bica2019estimating, seedat2022continuous}.

\textbf{Treatment plans.}
We differentiate between a sequential and concurrent treatment regime \citep{curran2011sequential}. In the sequential treatment arm, patients receive weekly chemotherapy for five weeks, followed by weekly radiotherapy for five weeks. In the concurrent treatment arm, patients biweekly receive both chemotherapy and radiotherapy for ten weeks. Patients are randomly divided among the two treatment arms based on a Bernoulli distribution with probability $p = 0.5$. This way, there is no confounding: treatment assignments are random and do not change during the trial.
\begin{figure*}[!t]
\centering
\includegraphics[width=\linewidth, trim={0 3pt 0 0}]{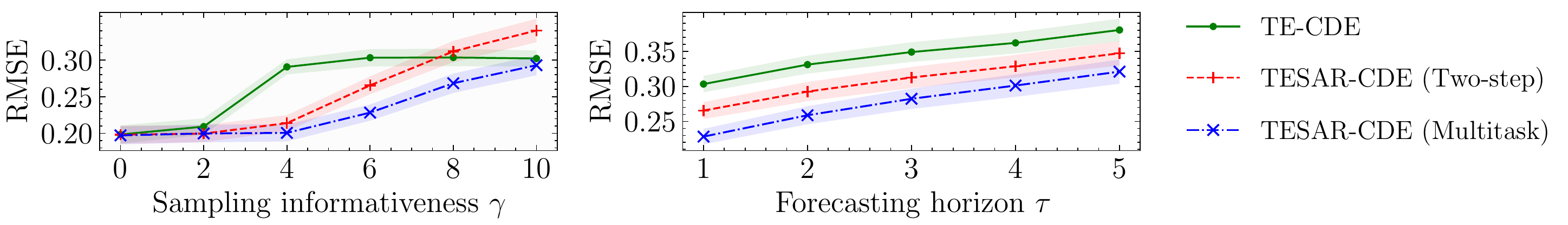}
\caption{\textbf{Results for varying informativeness $\gamma$ and different forecasting horizons $\tau$.} We show the RMSE $\pm$ SE over ten runs. \textbf{(Left)} RMSE for increasing levels of informativeness $\gamma$, keeping the forecasting horizon fixed at $\tau = 1$. \textbf{(Right)} RMSE for an increasing forecasting horizon $\tau$ up to five days, keeping informativeness fixed at $\gamma = 6$.}
\vspace{-8pt}
\label{fig:main_results}
\end{figure*}

\textbf{Observation process.}
We observe patients based on a patient-specific, history-dependent intensity process. This is achieved by simulating each patient's observation process with intensity $\lambda_i(t)$ in which $\gamma$ controls the informativeness:
\begin{equation}
    \lambda_i(t) = \sigma \left[ \gamma \left( \frac{\Bar{D}_i(t-)}{D_\text{max}} - \frac{1}{2} \right) \right],
\label{eq:obs_process}
\end{equation}
where $\sigma$ denotes the sigmoid function. $D_\text{max} = 13\text{cm}$ denotes the maximal tumor diameter and $\Bar{D}(t-)$ is the average tumor diameter over the past 15 days.
By simulating the observation process in this way, we can control the degree of informativeness: $\gamma=0$ implies SCAR as $\lambda_i(t) = 0.5$, while $\gamma>0$ implies SAR with a larger $\gamma$ implying more informativeness or dependence between the tumor size and intensity. As $\gamma$ increases, patients with larger tumors are more likely to be observed, those with smaller tumors less.

\textbf{Experimental setup.}
We assume treatments are always observed, as these are planned in advance and administered in the hospital. Nevertheless, at treatment time, we do not necessarily observe the patient's tumor size, e.g., because observing tumor size requires an invasive procedure separate from the treatment. For the test set, we observe all information at daily intervals. This idealized setup allows us to assess whether the model is able to learn the underlying distribution, as opposed to fitting the observed samples. Similarly, the test data also contains the potential outcomes for both treatment arms. For each experiment, we generate a training set with 200 patients, validation set with 50 patients, and test set with 200 patients, all over a period of 120 days.


\subsection{Empirical results}

This section presents the empirical results using the experimental setup described above. More specifically, we aim to answer three questions: (1) What is the impact of informative sampling?; (2) What is the impact of observation scarcity?; and (3) When does informativeness matter? In \cref{app:Additional_results}, we present additional experiments to evaluate TESAR-CDE's intensity prediction and analyze the sensitivity of the multitask configuration to hyperparameter $\alpha$.

\begin{figure}[t]
\centering
\includegraphics[width=\linewidth, trim={0 3pt 0 3pt}]{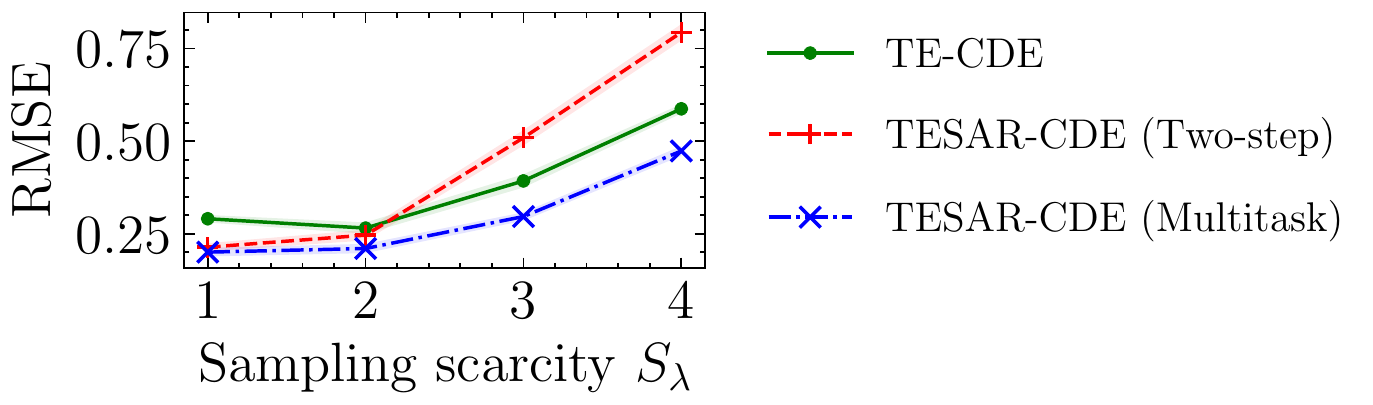}
\caption{\textbf{Observation scarcity.} We show the RMSE $\pm$ SE over ten runs at increasing observation scarcity $S_{\lambda}$ for fixed informativeness ($\gamma = 4$) and forecasting horizon ($\tau = 1$).}
\vspace{-2pt}
\rule{\textwidth}{0.75pt}
\vspace{-20pt}
\label{fig:max_intensities}
\end{figure}

\textbf{What is the impact of informative sampling?}
We compare the different models at varying levels of informativeness in \cref{fig:main_results}. Increasing $\gamma$ makes the observation process more informative by having patients with a large tumor visit more and patients with small tumors less (\cref{eq:obs_process}). The left side of \cref{fig:main_results} shows the RMSE of the different models at increasing informativeness. As sampling becomes more informative, TE-CDE's performance deteriorates and is outperformed by the proposed TESAR-CDE. The multitask variant in particular is robust to high levels of informativeness, achieving the lowest RMSE overall. The two-step variant generally outperforms TE-CDE, but performs worse for very high informativeness. As a high $\gamma$ results in very low intensities for some patients, the importance weights of these observations induce large variance and worse generalization properties in the weighted loss. This phenomenon is a well-known issue in importance weighting more generally \citep{cortes2010learning}. The right of \cref{fig:main_results} shows the performance at different forecasting horizons $\tau$ ranging from one to five days at a fixed informativeness of $\gamma = 6$. Both TESAR-CDE variants outperform the standard TE-CDE over all horizons, with the multitask variant again performing the best overall. 

\textbf{What is the impact of observation scarcity?} We analyze the influence of less frequent sampling across all patients. We simulate lower overall sampling by scaling all intensities as $\lambda'(t) =\frac{\lambda(t)}{S_\lambda}$ with ${S_\lambda} \in \{1, 2, 3, 4\}$. 
\cref{fig:max_intensities} shows the impact of increasing scarcity on the resulting RMSE. As expected, all models perform worse with less frequent sampling. The two-step TESAR-CDE performs worse than the baseline TE-CDE as scarcity increases, while the multitask TESAR-CDE is again the best performing model overall. This result indicates that the parameter efficiency of the multitask model can be helpful when sampling is scarce.

\begin{figure}[t]
\centering
\includegraphics[width=\linewidth, trim={0 3pt 0 3pt}]{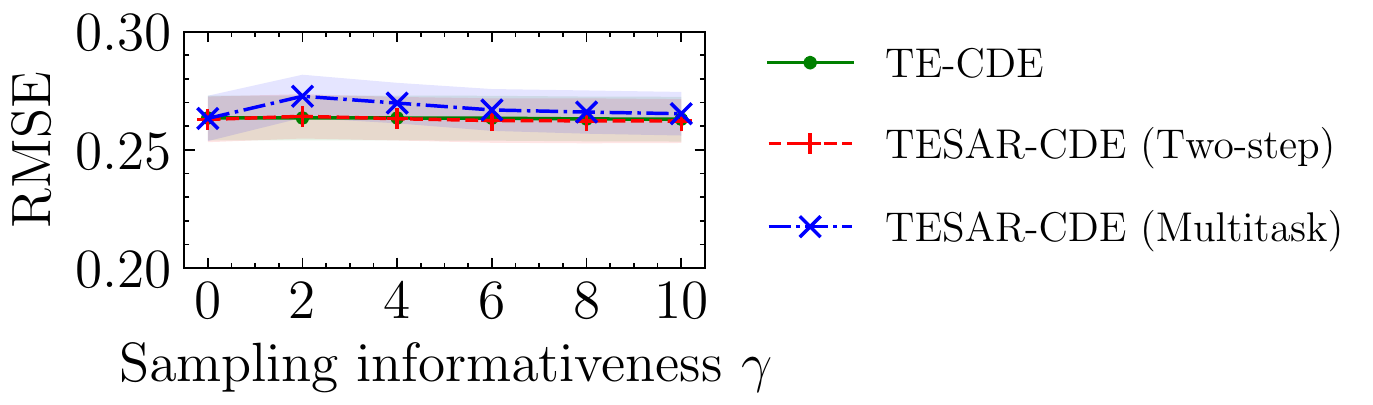}
\caption{\textbf{Outcome-unrelated sampling.} We show the RMSE $\pm$ SE over ten runs for a sampling mechanism unrelated to the outcome $Y(t)$ in function of informativeness $\gamma$.}
\label{fig:uniformative_sampling}
\vspace{-5pt} 
\end{figure}

\textbf{When does informativeness matter?}
The previous experiments analyzed informative observation processes where the observation intensity $\lambda(t)$ was directly related to the outcome $Y(t)$. Next, we analyze a special case of SAR where the intensity depends on information that is \textit{completely unrelated} to the outcome or treatments. This extreme scenario mimics a situation in which there are patients that visit often for reasons unrelated to underlying symptoms or outcome, e.g. when  they suffer from hypochondria. We examine this using an intensity that depends on covariates $\textstyle x^{\lambda}$: $\textstyle \lambda_i(t) = \sigma \left( \gamma \sum_j^d(c_j x_j) \right)$, where we include $d = 10$ static variables $x^{(i)}$ for each patient, with each $x_j \sim \mathcal{N}(0, 1)$ influencing the intensity through a coefficient $c_j \sim \mathcal{U}(-1, 1)$, but not affecting the patient in any other way. \cref{fig:uniformative_sampling} shows the prediction error of the different models for an increasing $\gamma$, averaged over $\tau \in \{1, \dots, 5\}$. In this scenario, the sampling mechanism does \textit{not} significantly affect performance, with all models having similar performance. If anything, TESAR-CDE (Multitask) performs slightly worse, possibly due to importance weighting being unnecessary and adding variance in this scenario. This result indicates that informative sampling may only matter when it depends on factors influencing \textit{both} observation intensity and outcome.

\section{Conclusion}

This work analyzed and formalized an essential challenge in learning to forecast treatment outcomes over time from observational data: the presence of informative sampling. We differentiated between different sampling mechanisms depending on the causal role of the observation intensity. This categorization allowed us to identify an overlooked, yet common setting in which observations are sampled irregularly over time with the observation intensity depending on the history of the instance's covariates, outcome, and/or treatments. We formalized learning in this context and analyzed it as a problem of covariate shift. Based on this, we proposed a general framework for learning under informative sampling and a novel method, TESAR-CDE, that instatiates this framework using Neural CDEs. Empirical results demonstrate the improved performance of TESAR-CDE over the current state-of-the-art when sampling is informative.


Accounting for informative sampling when learning to forecast treatment outcomes relies on strong identification assumptions regarding both the treatment assignment and sampling mechanism. As these assumptions are untestable, we need to rely on domain expertise to judge their plausibility in practical applications. For example, the sampling at random assumption would be violated if the observation intensity is affected by an unobserved cause of the outcome. While beyond the scope of the current work, we believe that exploring learning under violations of these assumptions is an important and fruitful area for future research -- similar to the rich lines of work exploring estimation of treatment effects with hidden confounders or missing treatment information.
{\parfillskip=0pt \emergencystretch=.5\textwidth \par}





\section*{Acknowledgements} We would like to thank Nabeel Seedat and the anonymous reviewers for insightful comments and discussions on earlier drafts of this paper. TV is supported by the Research Foundation -- Flanders (FWO PhD Fellowship 11I7322N). AC gratefully acknowledges funding from AstraZeneca.

\setlength{\bibsep}{7.6pt}
\bibliography{bibliography}

\begin{thebibliography}{69}
\providecommand{\natexlab}[1]{#1}
\providecommand{\url}[1]{\texttt{#1}}
\expandafter\ifx\csname urlstyle\endcsname\relax
  \providecommand{\doi}[1]{doi: #1}\else
  \providecommand{\doi}{doi: \begingroup \urlstyle{rm}\Url}\fi

\bibitem[Alaa \& Van Der~Schaar(2016)Alaa and Van
  Der~Schaar]{alaa2016balancing}
Alaa, A.~M. and Van Der~Schaar, M.
\newblock Balancing suspense and surprise: Timely decision making with
  endogenous information acquisition.
\newblock \emph{Advances in Neural Information Processing Systems}, 29, 2016.

\bibitem[Alaa et~al.(2017)Alaa, Hu, and Schaar]{alaa2017learning}
Alaa, A.~M., Hu, S., and Schaar, M.
\newblock Learning from clinical judgments: Semi-markov-modulated marked hawkes
  processes for risk prognosis.
\newblock In \emph{International Conference on Machine Learning}, pp.\  60--69.
  PMLR, 2017.

\bibitem[Aup{\'e}rin et~al.(2010)Aup{\'e}rin, Le~P{\'e}choux, Rolland, Curran,
  Furuse, Fournel, Belderbos, Clamon, Ulutin, Paulus, et~al.]{auperin2010meta}
Aup{\'e}rin, A., Le~P{\'e}choux, C., Rolland, E., Curran, W.~J., Furuse, K.,
  Fournel, P., Belderbos, J., Clamon, G., Ulutin, H.~C., Paulus, R., et~al.
\newblock Meta-analysis of concomitant versus sequential radiochemotherapy in
  locally advanced non-small-cell lung cancer.
\newblock \emph{Database of Abstracts of Reviews of Effects (DARE):
  Quality-assessed Reviews [Internet]}, 2010.

\bibitem[Bellot \& Van Der~Schaar(2021)Bellot and Van
  Der~Schaar]{bellot2021policy}
Bellot, A. and Van Der~Schaar, M.
\newblock Policy analysis using synthetic controls in continuous-time.
\newblock In \emph{International Conference on Machine Learning}, pp.\
  759--768. PMLR, 2021.

\bibitem[Berrevoets et~al.(2021)Berrevoets, Curth, Bica, McKinney, and van~der
  Schaar]{berrevoets2021disentangled}
Berrevoets, J., Curth, A., Bica, I., McKinney, E., and van~der Schaar, M.
\newblock Disentangled counterfactual recurrent networks for treatment effect
  inference over time.
\newblock \emph{arXiv preprint arXiv:2112.03811}, 2021.

\bibitem[Berrevoets et~al.(2023)Berrevoets, Imrie, Kyono, Jordon, and van~der
  Schaar]{berrevoets2023impute}
Berrevoets, J., Imrie, F., Kyono, T., Jordon, J., and van~der Schaar, M.
\newblock To impute or not to impute? missing data in treatment effect
  estimation.
\newblock In \emph{International Conference on Artificial Intelligence and
  Statistics}, pp.\  3568--3590. PMLR, 2023.

\bibitem[Bica et~al.(2019)Bica, Alaa, Jordon, and van~der
  Schaar]{bica2019estimating}
Bica, I., Alaa, A.~M., Jordon, J., and van~der Schaar, M.
\newblock Estimating counterfactual treatment outcomes over time through
  adversarially balanced representations.
\newblock In \emph{International Conference on Learning Representations}, 2019.

\bibitem[Biewald(2020)]{wandb}
Biewald, L.
\newblock Experiment tracking with weights and biases, 2020.
\newblock URL \url{https://www.wandb.com/}.
\newblock Software available from wandb.com.

\bibitem[Bu{\v{z}}kov{\'a} \& Lumley(2009)Bu{\v{z}}kov{\'a} and
  Lumley]{buuvzkova2009semiparametric}
Bu{\v{z}}kov{\'a}, P. and Lumley, T.
\newblock Semiparametric modeling of repeated measurements under
  outcome-dependent follow-up.
\newblock \emph{Statistics in Medicine}, 28\penalty0 (6):\penalty0 987--1003,
  2009.

\bibitem[Che et~al.(2018)Che, Purushotham, Cho, Sontag, and
  Liu]{che2018recurrent}
Che, Z., Purushotham, S., Cho, K., Sontag, D., and Liu, Y.
\newblock Recurrent neural networks for multivariate time series with missing
  values.
\newblock \emph{Scientific reports}, 8\penalty0 (1):\penalty0 1--12, 2018.

\bibitem[Chen et~al.(2018)Chen, Rubanova, Bettencourt, and
  Duvenaud]{chen2018neural}
Chen, R.~T., Rubanova, Y., Bettencourt, J., and Duvenaud, D.~K.
\newblock Neural ordinary differential equations.
\newblock \emph{Advances in neural information processing systems}, 31, 2018.

\bibitem[Clithero(2018)]{clithero2018response}
Clithero, J.~A.
\newblock Response times in economics: Looking through the lens of sequential
  sampling models.
\newblock \emph{Journal of Economic Psychology}, 69:\penalty0 61--86, 2018.

\bibitem[Cole \& Hern{\'a}n(2008)Cole and Hern{\'a}n]{cole2008constructing}
Cole, S.~R. and Hern{\'a}n, M.~A.
\newblock Constructing inverse probability weights for marginal structural
  models.
\newblock \emph{American journal of epidemiology}, 168\penalty0 (6):\penalty0
  656--664, 2008.

\bibitem[Cortes et~al.(2010)Cortes, Mansour, and Mohri]{cortes2010learning}
Cortes, C., Mansour, Y., and Mohri, M.
\newblock Learning bounds for importance weighting.
\newblock \emph{Advances in neural information processing systems}, 23, 2010.

\bibitem[Curran~Jr et~al.(2011)Curran~Jr, Paulus, Langer, Komaki, Lee, Hauser,
  Movsas, Wasserman, Rosenthal, Gore, et~al.]{curran2011sequential}
Curran~Jr, W.~J., Paulus, R., Langer, C.~J., Komaki, R., Lee, J.~S., Hauser,
  S., Movsas, B., Wasserman, T., Rosenthal, S.~A., Gore, E., et~al.
\newblock Sequential vs concurrent chemoradiation for stage iii non--small cell
  lung cancer: randomized phase iii trial rtog 9410.
\newblock \emph{Journal of the National Cancer Institute}, 103\penalty0
  (19):\penalty0 1452--1460, 2011.

\bibitem[Curth \& van~der Schaar(2021{\natexlab{a}})Curth and van~der
  Schaar]{curth2021inductive}
Curth, A. and van~der Schaar, M.
\newblock On inductive biases for heterogeneous treatment effect estimation.
\newblock \emph{Advances in Neural Information Processing Systems},
  34:\penalty0 15883--15894, 2021{\natexlab{a}}.

\bibitem[Curth \& van~der Schaar(2021{\natexlab{b}})Curth and van~der
  Schaar]{curth2021nonparametric}
Curth, A. and van~der Schaar, M.
\newblock Nonparametric estimation of heterogeneous treatment effects: From
  theory to learning algorithms.
\newblock In \emph{International Conference on Artificial Intelligence and
  Statistics}, pp.\  1810--1818. PMLR, 2021{\natexlab{b}}.

\bibitem[Curth \& van~der Schaar(2023)Curth and van~der
  Schaar]{curth2023understanding}
Curth, A. and van~der Schaar, M.
\newblock Understanding the impact of competing events on heterogeneous
  treatment effect estimation from time-to-event data.
\newblock In \emph{International Conference on Artificial Intelligence and
  Statistics}, pp.\  7961--7980. PMLR, 2023.

\bibitem[Curth et~al.(2021)Curth, Lee, and van~der Schaar]{curth2021survite}
Curth, A., Lee, C., and van~der Schaar, M.
\newblock Survite: Learning heterogeneous treatment effects from time-to-event
  data.
\newblock \emph{Advances in Neural Information Processing Systems},
  34:\penalty0 26740--26753, 2021.

\bibitem[De~Brouwer et~al.(2022)De~Brouwer, Gonzalez, and
  Hyland]{de2022predicting}
De~Brouwer, E., Gonzalez, J., and Hyland, S.
\newblock Predicting the impact of treatments over time with uncertainty aware
  neural differential equations.
\newblock In \emph{International Conference on Artificial Intelligence and
  Statistics}, pp.\  4705--4722. PMLR, 2022.

\bibitem[Del~Moral \& Murray(2015)Del~Moral and Murray]{del2015sequential}
Del~Moral, P. and Murray, L.~M.
\newblock Sequential monte carlo with highly informative observations.
\newblock \emph{SIAM/ASA Journal on Uncertainty Quantification}, 3\penalty0
  (1):\penalty0 969--997, 2015.

\bibitem[Farahani et~al.(2021)Farahani, Voghoei, Rasheed, and
  Arabnia]{farahani2021brief}
Farahani, A., Voghoei, S., Rasheed, K., and Arabnia, H.~R.
\newblock A brief review of domain adaptation.
\newblock \emph{Advances in data science and information engineering}, pp.\
  877--894, 2021.

\bibitem[Farzanfar et~al.(2017)Farzanfar, Abumuamar, Kim, Sirotich, Wang, and
  Pullenayegum]{farzanfar2017longitudinal}
Farzanfar, D., Abumuamar, A., Kim, J., Sirotich, E., Wang, Y., and
  Pullenayegum, E.
\newblock Longitudinal studies that use data collected as part of usual care
  risk reporting biased results: a systematic review.
\newblock \emph{BMC Medical Research Methodology}, 17\penalty0 (1):\penalty0
  1--12, 2017.

\bibitem[Furuse et~al.(1999)Furuse, Fukuoka, Kawahara, Nishikawa, Takada,
  Kudoh, Katagami, and Ariyoshi]{furuse1999phase}
Furuse, K., Fukuoka, M., Kawahara, M., Nishikawa, H., Takada, Y., Kudoh, S.,
  Katagami, N., and Ariyoshi, Y.
\newblock Phase iii study of concurrent versus sequential thoracic radiotherapy
  in combination with mitomycin, vindesine, and cisplatin in unresectable stage
  iii non--small-cell lung cancer.
\newblock \emph{Journal of Clinical Oncology}, 17\penalty0 (9):\penalty0
  2692--2692, 1999.

\bibitem[Gasparini et~al.(2020)Gasparini, Abrams, Barrett, Major, Sweeting,
  Brunskill, and Crowther]{gasparini2020mixed}
Gasparini, A., Abrams, K.~R., Barrett, J.~K., Major, R.~W., Sweeting, M.~J.,
  Brunskill, N.~J., and Crowther, M.~J.
\newblock Mixed-effects models for health care longitudinal data with an
  informative visiting process: A monte carlo simulation study.
\newblock \emph{Statistica Neerlandica}, 74\penalty0 (1):\penalty0 5--23, 2020.

\bibitem[Geng et~al.(2017)Geng, Paganetti, and Grassberger]{geng2017prediction}
Geng, C., Paganetti, H., and Grassberger, C.
\newblock Prediction of treatment response for combined chemo-and radiation
  therapy for non-small cell lung cancer patients using a bio-mathematical
  model.
\newblock \emph{Scientific reports}, 7\penalty0 (1):\penalty0 1--12, 2017.

\bibitem[Gische et~al.(2021)Gische, West, and Voelkle]{gische2021forecasting}
Gische, C., West, S.~G., and Voelkle, M.~C.
\newblock Forecasting causal effects of interventions versus predicting future
  outcomes.
\newblock \emph{Structural Equation Modeling: A Multidisciplinary Journal},
  28\penalty0 (3):\penalty0 475--492, 2021.

\bibitem[Goldstein et~al.(2016)Goldstein, Bhavsar, Phelan, and
  Pencina]{goldstein2016controlling}
Goldstein, B.~A., Bhavsar, N.~A., Phelan, M., and Pencina, M.~J.
\newblock Controlling for informed presence bias due to the number of health
  encounters in an electronic health record.
\newblock \emph{American journal of epidemiology}, 184\penalty0 (11):\penalty0
  847--855, 2016.

\bibitem[Gwak et~al.(2020)Gwak, Sim, Poli, Massaroli, Choo, and
  Choi]{gwak2020neural}
Gwak, D., Sim, G., Poli, M., Massaroli, S., Choo, J., and Choi, E.
\newblock Neural ordinary differential equations for intervention modeling.
\newblock \emph{arXiv preprint arXiv:2010.08304}, 2020.

\bibitem[Hassanpour \& Greiner(2019)Hassanpour and
  Greiner]{hassanpour2019counterfactual}
Hassanpour, N. and Greiner, R.
\newblock Counterfactual regression with importance sampling weights.
\newblock In \emph{IJCAI}, pp.\  5880--5887, 2019.

\bibitem[Hassanpour \& Greiner(2020)Hassanpour and
  Greiner]{hassanpour2019learning}
Hassanpour, N. and Greiner, R.
\newblock Learning disentangled representations for counterfactual regression.
\newblock In \emph{International Conference on Learning Representations}, 2020.

\bibitem[Hern{\'a}n et~al.(2009)Hern{\'a}n, McAdams, McGrath, Lanoy, and
  Costagliola]{hernan2009observation}
Hern{\'a}n, M.~A., McAdams, M., McGrath, N., Lanoy, E., and Costagliola, D.
\newblock Observation plans in longitudinal studies with time-varying
  treatments.
\newblock \emph{Statistical methods in medical research}, 18\penalty0
  (1):\penalty0 27--52, 2009.

\bibitem[Jeanselme et~al.(2022)Jeanselme, Martin, Peek, Sperrin, Tom, and
  Barrett]{jeanselme2022deepjoint}
Jeanselme, V., Martin, G., Peek, N., Sperrin, M., Tom, B., and Barrett, J.
\newblock Deepjoint: Robust survival modelling under clinical presence shift.
\newblock \emph{arXiv preprint arXiv:2205.13481}, 2022.

\bibitem[Johansson et~al.(2016)Johansson, Shalit, and
  Sontag]{johansson2016learning}
Johansson, F., Shalit, U., and Sontag, D.
\newblock Learning representations for counterfactual inference.
\newblock In \emph{International conference on machine learning}, pp.\
  3020--3029. PMLR, 2016.

\bibitem[Kennedy(2020)]{kennedy2020efficient}
Kennedy, E.~H.
\newblock Efficient nonparametric causal inference with missing exposure
  information.
\newblock \emph{The international journal of biostatistics}, 16\penalty0 (1),
  2020.

\bibitem[Kidger et~al.(2020)Kidger, Morrill, Foster, and
  Lyons]{kidger2020neural}
Kidger, P., Morrill, J., Foster, J., and Lyons, T.
\newblock Neural controlled differential equations for irregular time series.
\newblock \emph{Advances in Neural Information Processing Systems},
  33:\penalty0 6696--6707, 2020.

\bibitem[Li et~al.(2021)Li, Shahn, Li, Lu, Chakraborty, Sow, Ghalwash, and
  Lehman]{li2021gnet}
Li, R., Shahn, Z., Li, J., Lu, M., Chakraborty, P., Sow, D., Ghalwash, M., and
  Lehman, L.-w.~H.
\newblock G-net: a deep learning approach to g-computation for counterfactual
  outcome prediction under dynamic treatment regimes.
\newblock In \emph{Machine Learning for Health}, 2021.

\bibitem[Lim et~al.(2018)Lim, Alaa, and van~der Schaar]{lim2018rmsn}
Lim, B., Alaa, A., and van~der Schaar, M.
\newblock Forecasting treatment responses over time using recurrent marginal
  structural networks.
\newblock In Bengio, S., Wallach, H., Larochelle, H., Grauman, K.,
  Cesa-Bianchi, N., and Garnett, R. (eds.), \emph{Advances in Neural
  Information Processing Systems}, volume~31. Curran Associates, Inc., 2018.
\newblock URL
  \url{https://proceedings.neurips.cc/paper/2018/file/56e6a93212e4482d99c84a639d254b67-Paper.pdf}.

\bibitem[Lin et~al.(2004)Lin, Scharfstein, and Rosenheck]{lin2004analysis}
Lin, H., Scharfstein, D.~O., and Rosenheck, R.~A.
\newblock Analysis of longitudinal data with irregular, outcome-dependent
  follow-up.
\newblock \emph{Journal of the Royal Statistical Society: Series B (Statistical
  Methodology)}, 66\penalty0 (3):\penalty0 791--813, 2004.

\bibitem[Little \& Rubin(2019)Little and Rubin]{little2019statistical}
Little, R.~J. and Rubin, D.~B.
\newblock \emph{Statistical analysis with missing data}, volume 793.
\newblock John Wiley \& Sons, 2019.

\bibitem[Liu et~al.(2008)Liu, Huang, and O'Quigley]{liu2008analysis}
Liu, L., Huang, X., and O'Quigley, J.
\newblock Analysis of longitudinal data in the presence of informative
  observational times and a dependent terminal event, with application to
  medical cost data.
\newblock \emph{Biometrics}, 64\penalty0 (3):\penalty0 950--958, 2008.

\bibitem[Lok(2008)]{lok2008statistical}
Lok, J.~J.
\newblock Statistical modeling of causal effects in continuous time.
\newblock \emph{The Annals of Statistics}, 36\penalty0 (3):\penalty0
  1464--1507, 2008.

\bibitem[Maia~Polo \& Vicente(2022)Maia~Polo and Vicente]{maia2022effective}
Maia~Polo, F. and Vicente, R.
\newblock Effective sample size, dimensionality, and generalization in
  covariate shift adaptation.
\newblock \emph{Neural Computing and Applications}, pp.\  1--13, 2022.

\bibitem[Mayer et~al.(2020)Mayer, Sverdrup, Gauss, Moyer, Wager, and
  Josse]{mayer2020doubly}
Mayer, I., Sverdrup, E., Gauss, T., Moyer, J.-D., Wager, S., and Josse, J.
\newblock Doubly robust treatment effect estimation with missing attributes.
\newblock \emph{Annals of Applied Statistics}, 14\penalty0 (3):\penalty0
  1409--1431, 2020.

\bibitem[McCulloch et~al.(2016)McCulloch, Neuhaus, and
  Olin]{mcculloch2016biased}
McCulloch, C.~E., Neuhaus, J.~M., and Olin, R.~L.
\newblock Biased and unbiased estimation in longitudinal studies with
  informative visit processes.
\newblock \emph{Biometrics}, 72\penalty0 (4):\penalty0 1315--1324, 2016.

\bibitem[Melnychuk et~al.(2022)Melnychuk, Frauen, and
  Feuerriegel]{melnychuk2022causaltransf}
Melnychuk, V., Frauen, D., and Feuerriegel, S.
\newblock Causal transformer for estimating counterfactual outcomes.
\newblock In Chaudhuri, K., Jegelka, S., Song, L., Szepesvari, C., Niu, G., and
  Sabato, S. (eds.), \emph{Proceedings of the 39th International Conference on
  Machine Learning}, volume 162 of \emph{Proceedings of Machine Learning
  Research}, pp.\  15293--15329. PMLR, 17--23 Jul 2022.
\newblock URL \url{https://proceedings.mlr.press/v162/melnychuk22a.html}.

\bibitem[Morrill et~al.(2021)Morrill, Kidger, Yang, and
  Lyons]{morrill2021neural}
Morrill, J., Kidger, P., Yang, L., and Lyons, T.
\newblock Neural controlled differential equations for online prediction tasks.
\newblock \emph{arXiv preprint arXiv:2106.11028}, 2021.

\bibitem[Pullenayegum \& Lim(2016)Pullenayegum and
  Lim]{pullenayegum2016longitudinal}
Pullenayegum, E.~M. and Lim, L.~S.
\newblock Longitudinal data subject to irregular observation: A review of
  methods with a focus on visit processes, assumptions, and study design.
\newblock \emph{Statistical methods in medical research}, 25\penalty0
  (6):\penalty0 2992--3014, 2016.

\bibitem[Qian et~al.(2021)Qian, Zhang, Bica, Wood, and van~der
  Schaar]{qian2021synctwin}
Qian, Z., Zhang, Y., Bica, I., Wood, A., and van~der Schaar, M.
\newblock Synctwin: Treatment effect estimation with longitudinal outcomes.
\newblock \emph{Advances in Neural Information Processing Systems},
  34:\penalty0 3178--3190, 2021.

\bibitem[Redko et~al.(2020)Redko, Morvant, Habrard, Sebban, and
  Bennani]{redko2020survey}
Redko, I., Morvant, E., Habrard, A., Sebban, M., and Bennani, Y.
\newblock A survey on domain adaptation theory: learning bounds and theoretical
  guarantees.
\newblock \emph{arXiv preprint arXiv:2004.11829}, 2020.

\bibitem[Robins(1986)]{robins1986new}
Robins, J.
\newblock A new approach to causal inference in mortality studies with a
  sustained exposure period—application to control of the healthy worker
  survivor effect.
\newblock \emph{Mathematical modelling}, 7\penalty0 (9-12):\penalty0
  1393--1512, 1986.

\bibitem[Robins(1997)]{robins1997causal}
Robins, J.~M.
\newblock Causal inference from complex longitudinal data.
\newblock In \emph{Latent variable modeling and applications to causality},
  pp.\  69--117. Springer, 1997.

\bibitem[Robins et~al.(1995)Robins, Rotnitzky, and Zhao]{robins1995analysis}
Robins, J.~M., Rotnitzky, A., and Zhao, L.~P.
\newblock Analysis of semiparametric regression models for repeated outcomes in
  the presence of missing data.
\newblock \emph{Journal of the american statistical association}, 90\penalty0
  (429):\penalty0 106--121, 1995.

\bibitem[Robins et~al.(2000)Robins, Hernan, and Brumback]{robins2000marginal}
Robins, J.~M., Hernan, M.~A., and Brumback, B.
\newblock Marginal structural models and causal inference in epidemiology,
  2000.

\bibitem[Roy et~al.(2017)Roy, Lum, and Daniels]{roy2017bayesian}
Roy, J., Lum, K.~J., and Daniels, M.~J.
\newblock A bayesian nonparametric approach to marginal structural models for
  point treatments and a continuous or survival outcome.
\newblock \emph{Biostatistics}, 18\penalty0 (1):\penalty0 32--47, 2017.

\bibitem[Rubanova et~al.(2019)Rubanova, Chen, and Duvenaud]{rubanova2019latent}
Rubanova, Y., Chen, R.~T., and Duvenaud, D.~K.
\newblock Latent ordinary differential equations for irregularly-sampled time
  series.
\newblock \emph{Advances in neural information processing systems}, 32, 2019.

\bibitem[Rubin(1976)]{rubin1976inference}
Rubin, D.~B.
\newblock Inference and missing data.
\newblock \emph{Biometrika}, 63\penalty0 (3):\penalty0 581--592, 1976.

\bibitem[Rubin(2005)]{rubin2005causal}
Rubin, D.~B.
\newblock Causal inference using potential outcomes: Design, modeling,
  decisions.
\newblock \emph{Journal of the American Statistical Association}, 100\penalty0
  (469):\penalty0 322--331, 2005.

\bibitem[Schulam \& Saria(2017)Schulam and Saria]{schulam2017reliable}
Schulam, P. and Saria, S.
\newblock Reliable decision support using counterfactual models.
\newblock \emph{Advances in neural information processing systems}, 30, 2017.

\bibitem[Seedat et~al.(2022)Seedat, Imrie, Bellot, Qian, and van~der
  Schaar]{seedat2022continuous}
Seedat, N., Imrie, F., Bellot, A., Qian, Z., and van~der Schaar, M.
\newblock Continuous-time modeling of counterfactual outcomes using neural
  controlled differential equations.
\newblock In \emph{International Conference on Machine Learning}, pp.\
  19497--19521. PMLR, 2022.

\bibitem[Shalit et~al.(2017)Shalit, Johansson, and
  Sontag]{shalit2017estimating}
Shalit, U., Johansson, F.~D., and Sontag, D.
\newblock Estimating individual treatment effect: generalization bounds and
  algorithms.
\newblock In \emph{International Conference on Machine Learning}, pp.\
  3076--3085. PMLR, 2017.

\bibitem[Shchur et~al.(2021)Shchur, T{\"u}rkmen, Januschowski, and
  G{\"u}nnemann]{shchur2021neural}
Shchur, O., T{\"u}rkmen, A.~C., Januschowski, T., and G{\"u}nnemann, S.
\newblock Neural temporal point processes: A review.
\newblock \emph{arXiv preprint arXiv:2104.03528}, 2021.

\bibitem[Shimodaira(2000)]{shimodaira2000improving}
Shimodaira, H.
\newblock Improving predictive inference under covariate shift by weighting the
  log-likelihood function.
\newblock \emph{Journal of statistical planning and inference}, 90\penalty0
  (2):\penalty0 227--244, 2000.

\bibitem[Soleimani et~al.(2017)Soleimani, Subbaswamy, and
  Saria]{soleimani2017treatment}
Soleimani, H., Subbaswamy, A., and Saria, S.
\newblock Treatment-response models for counterfactual reasoning with
  continuous-time, continuous-valued interventions.
\newblock \emph{arXiv preprint arXiv:1704.02038}, 2017.

\bibitem[Vanderschueren et~al.(2023)Vanderschueren, Boute, Verdonck, Baesens,
  and Verbeke]{vanderschueren2023optimizing}
Vanderschueren, T., Boute, R., Verdonck, T., Baesens, B., and Verbeke, W.
\newblock Optimizing the preventive maintenance frequency with causal machine
  learning.
\newblock \emph{International Journal of Production Economics}, 258:\penalty0
  108798, 2023.

\bibitem[VanderWeele(2019)]{vanderweele2019principles}
VanderWeele, T.~J.
\newblock Principles of confounder selection.
\newblock \emph{European journal of epidemiology}, 34:\penalty0 211--219, 2019.

\bibitem[Xu et~al.(2016)Xu, Xu, and Saria]{xu2016bayesian}
Xu, Y., Xu, Y., and Saria, S.
\newblock A bayesian nonparametric approach for estimating individualized
  treatment-response curves.
\newblock In \emph{Machine learning for healthcare conference}, pp.\  282--300.
  PMLR, 2016.

\bibitem[Yoon et~al.(2018)Yoon, Zame, and Van Der~Schaar]{yoon2018deep}
Yoon, J., Zame, W.~R., and Van Der~Schaar, M.
\newblock Deep sensing: Active sensing using multi-directional recurrent neural
  networks.
\newblock In \emph{International Conference on Learning Representations}, 2018.

\bibitem[Yu et~al.(2009)Yu, Krishnapuram, Rosales, and Rao]{yu2009active}
Yu, S., Krishnapuram, B., Rosales, R., and Rao, R.~B.
\newblock Active sensing.
\newblock In \emph{Artificial Intelligence and Statistics}, pp.\  639--646.
  PMLR, 2009.

\end{thebibliography}
\bibliographystyle{icml2023}

\newpage
\appendix
\onecolumn
\section{Extended Related Work}\label{sec:appendix_related_work}

This section provides a more extensive discussion of several related areas of work.

\subsection{Forecasting treatment effects over time}

There is a growing interest in the ML literature in estimating personalized treatment effects over time. To this aim, different types of neural networks have been explored, including RNNs \citep{lim2018rmsn, bica2019estimating, li2021gnet, berrevoets2021disentangled}, transformers \citep{melnychuk2022causaltransf}, and Neural ODEs \citep{gwak2020neural,de2022predicting} or Neural CDEs \citep{seedat2022continuous}. All existing work in this area has implicitly relied upon assumptions of the observation process, assuming regular observations or completely random observation intervals, see \cref{tab:lit_table}. Conversely, our work relies on the less strict SAR assumption.

\begin{table}[h]
\centering
\small
\bgroup
\def\arraystretch{0.75}
\setlength{\tabcolsep}{8pt}
\begin{tabular}{L{100pt}L{50pt}}
\toprule
   \textbf{Reference} & \textbf{Sampling} \\
\midrule
   \citet{lim2018rmsn} & Regular \\
   \citet{bica2019estimating} & Regular \\
   \citet{berrevoets2021disentangled} & Regular \\
   \citet{li2021gnet} & Regular \\
   \citet{melnychuk2022causaltransf} & Regular \\
   \midrule
   \citet{gwak2020neural} & SCAR \\
   \citet{seedat2022continuous} & SCAR \\
   \citet{de2022predicting} & SCAR \\
   \midrule
   This work & SAR \\
\bottomrule
\end{tabular}
\egroup
\caption{\textbf{An overview of existing work.} We categorize the related work according to the assumptions made regarding the observation process. Sampling is either assumed to be regular, completely at random (SCAR), or at random (SAR).}
\label{tab:lit_table}
\end{table}

In addition to existing work leveraging neural networks, there is another line of work in the ML literature that uses Gaussian processes to estimate treatment effects over time in the presence of irregular samples \citep{xu2016bayesian, schulam2017reliable, roy2017bayesian, soleimani2017treatment}. Other work has looked at using synthetic controls to estimate the effect of a (single) intervention over time \citep{bellot2021policy, qian2021synctwin}. Similar to the ML literature on estimating treatment effects over time using neural networks discussed above, these also rely on regular samples or SCAR. To the best of our knowledge, no existing work in the ML literature has studied the problem of estimating treatment effects given SAR, which is the focus of this work.

Outside the ML literature, several approaches have been proposed for estimating \textit{average} treatment effects over time, most notably the seminal work using $g$-computation and marginal structural models \citep{robins1986new, robins1997causal, robins2000marginal}. More specifically, several approaches have been proposed in the (bio)statistics literature to learn causal effects under informative sampling, see \citet{gasparini2020mixed} for an overview. Existing work on estimating causal effects under SAR can be categorized based on (1) whether they use inverse visiting weights or random effects, and (2) whether they assume discrete or continuous time. (1) The first group uses the \textit{inverse probability of visiting} or its continuous-time equivalent, the inverse intensity of visiting, as weights in the objective function of the estimator \citep{robins1995analysis, lin2004analysis, pullenayegum2016longitudinal}. (2) The second uses \textit{shared random effects} to jointly model the observation and outcome processes \citep{liu2008analysis}. However, these approaches assume a certain parametric form or latent variable(s), which might not reflect the actual (unknown) data generating procedure and usually focus on population average effects. Conversely, our approach is conceptually similar to inverse intensity of visit weighting, but leverages the use of flexible ML methods that do not require these assumptions.

\subsection{Informative sampling in ML}
Various other works in ML consider related problem settings. For example, informative sampling has been considered as a source of information for prognosis in health care \citep{alaa2017learning} and as a challenge to robustness of predictive models to distribution shifts \cite{jeanselme2022deepjoint}, though this line of work does not consider the estimation of treatment effects. Moreover, the literature on active sensing views takes observing as an active role in which the decision-maker controls the sampling mechanism \citep{yu2009active}. The key question addressed in this line of work is what and when to measure \citep{yoon2018deep} and, potentially, also when to stop measuring \citep{alaa2016balancing}. This is in contrast to our setting, where data is not actively sampled, but \textit{observed} over time and the observer has a passive role.

\subsection{Neural ODEs}
Neural ordinary differential equations (ODEs) have recently emerged as a novel class of machine learning models combining neural networks and differential equations \citep{chen2018neural}. Due to their ability of handling irregular observations, Neural ODEs have been applied for time series, either directly or combined with a recurrent neural network \citep[e.g., ][]{rubanova2019latent}. Neural controlled differential equations (CDEs) additionally allow for modeling covariates as a control, making them suitable for dealing with irregularly sampled time series \citep{kidger2020neural, morrill2021neural}, as in the setting considered in this work.

\subsection{Missing data}
Dealing with missing data is an important and established field in statistics and machine learning \citep{rubin1976inference, little2019statistical}. This literature is related to our setting, as we are interested in learning a continuous latent path based on irregular observations over time, which could also be seen as a form of missing data imputation. Moreover, the sampling mechanisms considered in this work are similar to missing data mechanisms. Several recent works explore dealing with missing data in the context of treatment effect estimation \citep{mayer2020doubly, berrevoets2023impute}. 

\clearpage
\section{List of Mathematical Symbols}

We compile a list of mathematical symbols and their explanation in \cref{tab:symbols}. Moreover, we use a real-world example of a health care application to illustrate their meaning.

\begin{table}[!h]
    \centering
    \setlength{\tabcolsep}{3.5pt}
    \begin{tabular}{rll} 
    \toprule
        \textbf{Symbol} & \textbf{Explanation} & \textbf{Cancer patient example} \\
    \midrule
        $X$ & Covariate path & Blood pressure, heart rate, etc. \\
        $A$ & Treatment path & Chemotherapy, radiotherapy, etc. \\
        $Y$ & Outcome path & Tumor size \\
        $t$ & Time & \\
        $N(t)$ & Counting process over time & Five observations after two weeks \\
        $\lambda(t)$ & Observation intensity over time & Probability of observing tumor size at time $t$ \\
        $\mu_{a,t}(\tau)$ & \hangindent=1em Expected treatment outcome at time $t+\tau$ given treatment path $a$ & \hangindent=1em Tumor size next week absent any treatment \\
        $\hat{y}_{i,t}(t')$ & Instance $i$'s predicted treatment outcome at time $t' = t+\tau$ & Patient $i$'s tumor size next week \\
        $\hat{\lambda}_{i,t}(t')$ & Instance $i$'s predicted observation intensity at time $t' = t+\tau$ & Patient $i$'s observation intensity next week \\
    \bottomrule
    \end{tabular}
    \caption{\textbf{List of symbols.} We compile a list of the main mathematical symbols used and their explanation. The final column illustrates each symbol for the case of a cancer patient.}
    \label{tab:symbols}
\end{table}

\clearpage
\section{TESAR-CDE: Multitask Training Procedure}
\label{sec:appendix_training}

We include a more detailed training procedure for the multitask configuration of TESAR-CDE in \cref{alg:tesar_cde_training}.
\begin{algorithm}
\caption{Pseudo-code for the TESAR-CDE (Multitask) training procedure}\label{alg:tesar_cde_training}
\begin{algorithmic}
    \STATE \textbf{Input:} Observational data $\mathcal{D} = \{t_j^{(i)}, x_{t_j}^{(i)}, a_{t_j}^{(i)}, y_{t_j}^{(i)}\}$ for $i \in \{0, \dots, n\}$ and $j \in \{0, \dots, m_i\}$, weighted MSE loss $\mathcal{L}^{\text{WMSE}}$, cross-entropy loss $\mathcal{L}^{\text{CE}}$, total epochs $E$, learning rate $\eta$, and batch size $b$. TESAR-CDE architecture consisting of four networks: an embedding network $g$ with weights $W_g$, an encoder CDE function $f_\theta$ with weights $W_\theta$, a decoder CDE function $f_\phi$ with weights $W_\phi$, a final intensity map $f_\psi^\lambda$ with weights $W_\psi^\lambda$, and a final outcome map $f_\psi^y$ with weights $W_\psi^y$. 
    \FOR{epochs = $1$ to $E$}
        \STATE Sample batch $i_0, i_1, \dots, i_b \subset \{0, \dots, n\}$
        \STATE Encode the first observation $z(t_0)^{(i)} = g(t_0^{(i)}, x_{t_0}^{(i)}, a_{t_0}^{(i)}, y_{t_0}^{(i)})$ for each $i$ in batch
        \STATE Encode the history up to time $t$: $z(t) = \texttt{ODESolve}(f_\theta, z(t_0), \Bar{X}(t), \Bar{A}(t), \Bar{Y}(t))$
        \STATE Decode the history up to time $t+\tau$: $z_t(t+\tau) = \texttt{ODESolve}(f_\theta, z(t), \Bar{A}_t(t+\tau))$
        \STATE Map to forecast the outcome at $t+\tau$: $\bar{y}_t(t+\tau) = f_\psi^y (z_t(t+\tau))$
        \STATE Map to forecast the intensity at $t+\tau$: $\bar{\lambda}_t(t+\tau) = f_\psi^\lambda (z_t(t+\tau))$
        \STATE Compute $\text{grad}_g = \nabla_{W_g}  \frac{1}{n} \sum_i^n\mathcal{L}^{\text{WMSE}}_i$
        \STATE Compute $\text{grad}_\theta = \nabla_{W_\theta}  \frac{1}{n} \sum_i^n \mathcal{L}^{\text{WMSE}}_i$
        \STATE Compute $\text{grad}_{\phi} = \nabla_{W_{\phi}}  \frac{1}{n} \sum_i^n \mathcal{L}^{\text{WMSE}}_i$
        \STATE Compute $\text{grad}_{\psi^y} = \nabla_{W_{\psi^y}}  \frac{1}{n} \sum_i^n \mathcal{L}^{\text{WMSE}}_i$
        \STATE Compute $\text{grad}_{\psi^\lambda} = \nabla_{W_{\psi^\lambda}} \frac{1}{n} \sum_i^n \mathcal{L}^{\text{CE}}_i$
        \STATE Update weights $W_{g} \leftarrow W_{g} - \eta \; \text{grad}_g$
        \STATE Update weights $W_{\theta} \leftarrow W_{\theta} - \eta \; \text{grad}_\theta$
        \STATE Update weights $W_{\phi} \leftarrow W_{\phi} - \eta \; \text{grad}_\phi$
        \STATE Update weights $W_\psi^y \leftarrow W_\psi^y - \eta \; \text{grad}_\phi$
        \STATE Update weights $W_\psi^\lambda \leftarrow W_\psi^\lambda - \eta \; \text{grad}_\psi^\lambda$
        \IF{If $\frac{1}{n} \sum_i^n \mathcal{L}^\text{MT}_i = \frac{1}{n} \sum_i^n \left( (1 - \alpha) \mathcal{L}^{\text{WMSE}}_i + \alpha \mathcal{L}^{\text{CE}}_i \right)$ did not improve for 50 epochs}
            \STATE Break \COMMENT{Early stopping}
        \ENDIF
    \ENDFOR
\end{algorithmic}
\end{algorithm}

\clearpage
\section{TESAR-CDE: Implementation}\label{sec:appendix_tesar_cde_implement}

This section provides more details on the implementation of TESAR-CDE.

\subsection{Weight truncation}
For more stable training, we truncate the estimated intensities at $c_\text{min}=0.001$, such that the maximal importance weight is equal to $1000$. This is similar to what is typically done with propensity scores when adjusting for confounding bias. The truncation constant $c_\text{min}$ allows for trading off bias and variance, with $c_\text{min}=1$ corresponding to the unweighted variant \citep{cole2008constructing}. We did not tune the cutoff rate $c_\text{min}$.

\subsection{Hyperparameter optimization}
To allow for a fair comparison between the models, we do not tune hyperparameters for each model separately, but rather find the best configuration for the baseline TE-CDE model at a level of informativeness $\gamma=0$ and use this for all models. We show the ranges and final values for all hyperparameters in \cref{tab:app_hyperparameter_opt}. Hyperparameter optimization was done using wandb's Bayesian optimization \citep{wandb}. For each network in the CDE ($f_\theta$ and $f_\phi$), we use a final tanh activation layer, as recommended by \citet{kidger2020neural}. All models are trained with a batch size of $128$ and learning rate of $5e-4$ for a maximum of $1000$ epochs. Learning was terminated if the training loss did not improve for $50$ epochs. For the multitask configuration, we use $\alpha = 0.8$ to balance the loss terms, though this is only used for early stopping as each part of the network has a different optimizer, see also \cref{alg:tesar_cde_training}. For all models, we construct a control path for the Neural CDEs using a cubic interpolation.

\begin{table}[!h]
\centering
\begin{tabular}{ll}
    \toprule
    \textbf{Parameter} & \textbf{Range} \\
    \midrule
    Latent state $z$ dimension & $\{8, 16, \mathbf{32}\}$ \\
    \midrule
    Encoder layers & $\{1, 2, \mathbf{3}\}$ \\
    Decoder layers & $\{1, \mathbf{2}, 3\}$ \\
    Map layers & $\{\mathbf{1}, 2\}$ \\
    \midrule
    Encoder hidden neurons & $\{4, \mathbf{8}, 16\}$ \\
    Decoder hidden neurons & $\{4, \mathbf{8}, 16\}$ \\
    Map hidden neurons & $\{4, \mathbf{8}, 16\}$ \\
    \bottomrule
\end{tabular}
\caption{\textbf{Hyperparameter optimization.} We show the range for each hyperparameter that was optimized. The optimal value is shown in {bold}.}
\label{tab:app_hyperparameter_opt}
\end{table}

\subsection{A note on adding counts}
A frequent practice in time series forecasting when observation times may be informative is to add observation counts to the data \citep{che2018recurrent, kidger2020neural}. However, in the context of estimating treatment effects, this is problematic. First, adding counts is complicated because estimating counterfactual treatments would then require counterfactual count data, which is not observed. Moreover, adding count data may itself introduce confounding or collider bias \citep{goldstein2016controlling}. Therefore, we do not add observation counts in this work.

\clearpage
\section{Tumor Growth Simulation}
\label{sec:app_simulation}

We use the tumor growth simulation of \citet{geng2017prediction}, which was also used in the previous ML literature on estamating treatment effects over time \citep[e.g., ][]{lim2018rmsn, bica2019estimating, melnychuk2022causaltransf, seedat2022continuous}. We refer to these works for more details.

The tumor size is modelled as:
\begin{align}
    \frac{dY(t)}{dt} = \Big[ 1  + \underbrace{\rho \log \left(\frac{K}{Y(t)}\right)}_{\text{Tumor growth}} - & \underbrace{\beta_c C_t}_{\text{Chemotherapy}} - \underbrace{\left(\alpha_r d(t) + \beta_r d(t)^2 \right)}_{\text{Radiotherapy}} + \underbrace{\epsilon_t}_{\text{Noise}} \Big] Y(t).
\notag
\end{align}
Parameters are obtained as follows. Carrying capacity $K$ is set equal to $30$. Growth parameter $\rho$ is sampled from a normal distribution $\rho \sim \mathcal{N}(7.00 \times 10^{-5}, 7.23 \times 10^{-3})$. $\beta_c$ is also sampled from a normal distribution $\beta_c \sim \mathcal{N}(0.028, 0.0007)$. Finally, $\alpha_r$ and $\beta_r$ are obtained as $\alpha_r \sim \mathcal{N}(0.0398, 0.168)$ and $\beta = \frac{\alpha}{10}$. Noise is added by sampling $\epsilon(t) \sim \mathcal{N}(0, 0.01)$.

Following earlier work \citep{lim2018rmsn, bica2019estimating, seedat2022continuous}, we create heterogeniety in the treatment effects by creating three patient groups. Patient group 1 has a larger radiotherapy effect, achieved by multiplying $\mu(\alpha_r)$ with 1.1. Similarly, patient group 3 has a larger chemotherapy effect by increasing $\alpha_c$ with 10\%.

\begin{figure}[h]
\centering
\begin{subfigure}[b]{0.4\linewidth}
    \centering
    \includegraphics[width=\linewidth]{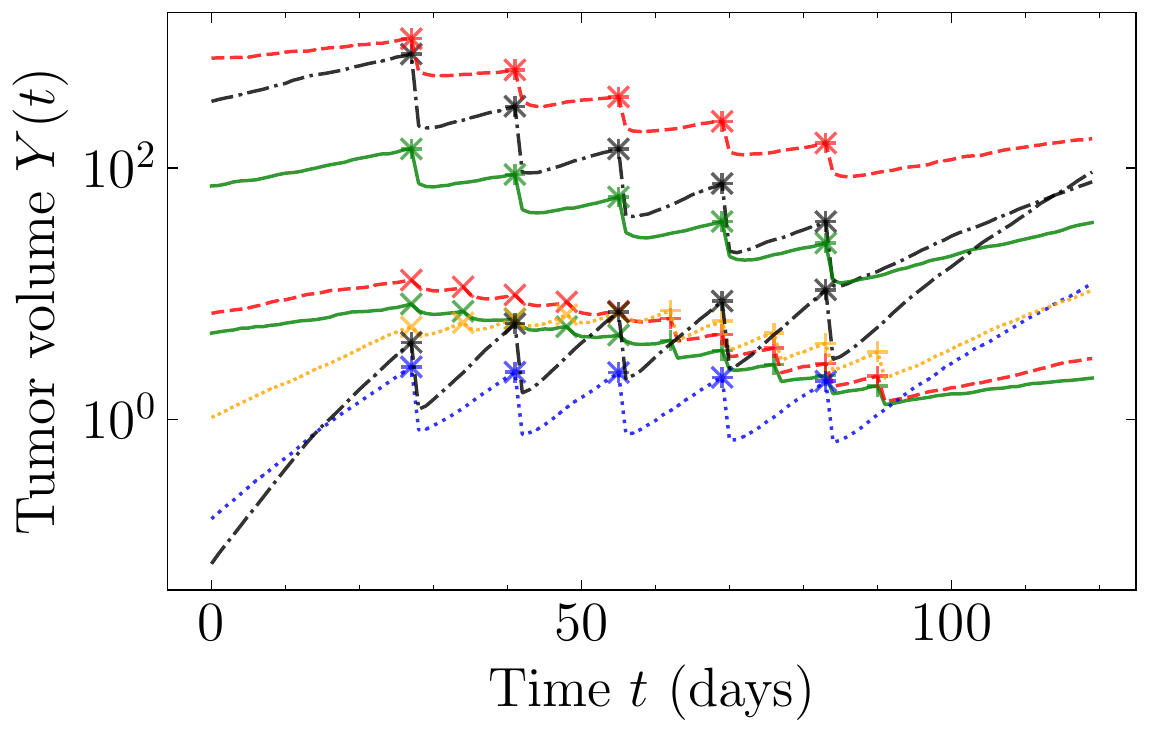}
    \caption{Tumor size $Y(t)$ over time}
\end{subfigure}
\hspace{50pt}
\begin{subfigure}[b]{0.4\linewidth}
        \centering
    \includegraphics[width=\linewidth]{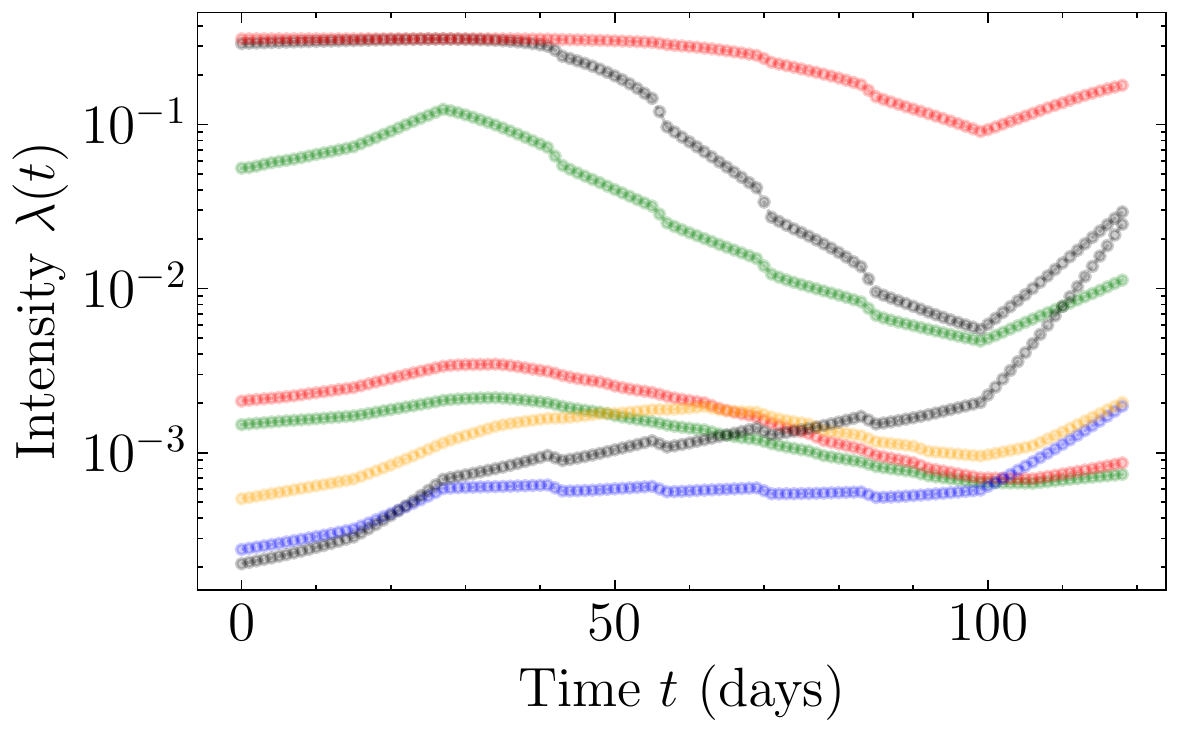}
    \caption{Intensity $\lambda(t)$ over time}
\end{subfigure}
\caption{\textbf{Tumor and intensity evolution.} We show the simulated tumor size and corresponding intensity over time for several (randomly selected) patients. The intensity is simulated is based on \cref{eq:obs_process} with an informativeness $\gamma = 4$.}
\label{fig:simulation}
\end{figure}

\begin{figure}[h]
\centering
\begin{subfigure}[b]{0.18\linewidth}
    \centering
    \includegraphics[width=\linewidth]{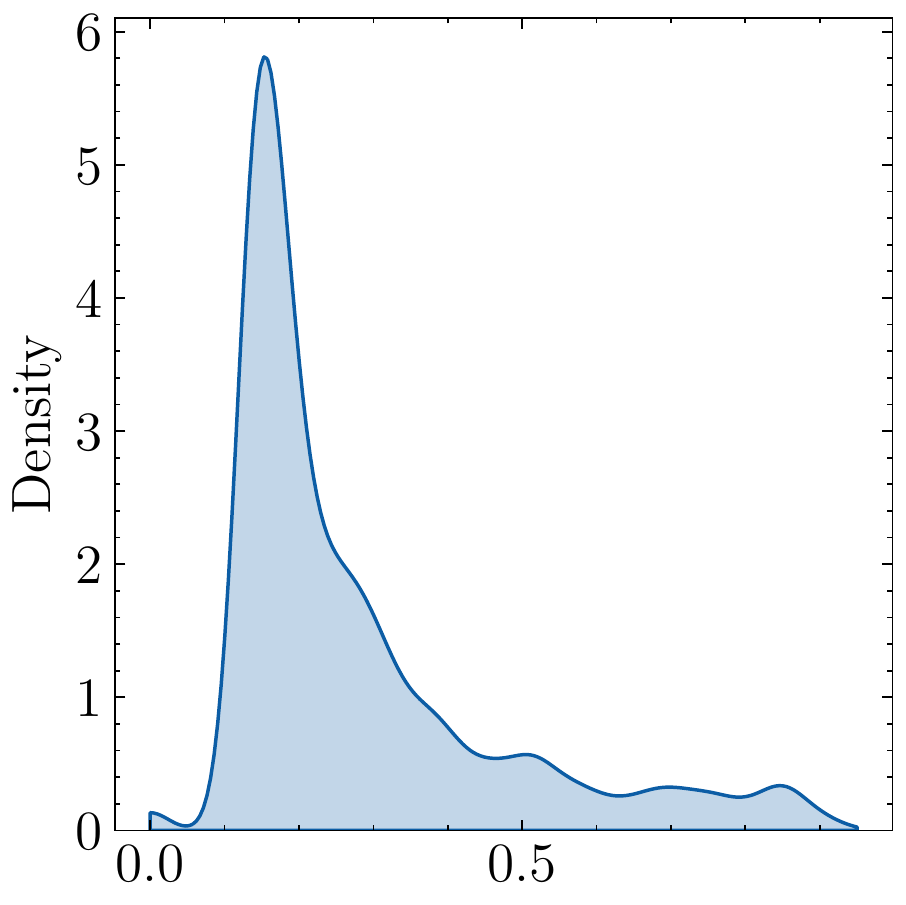}
    \caption{$\gamma = 2$}
\end{subfigure}
\hfill
\begin{subfigure}[b]{0.18\linewidth}
    \centering
    \includegraphics[width=\linewidth]{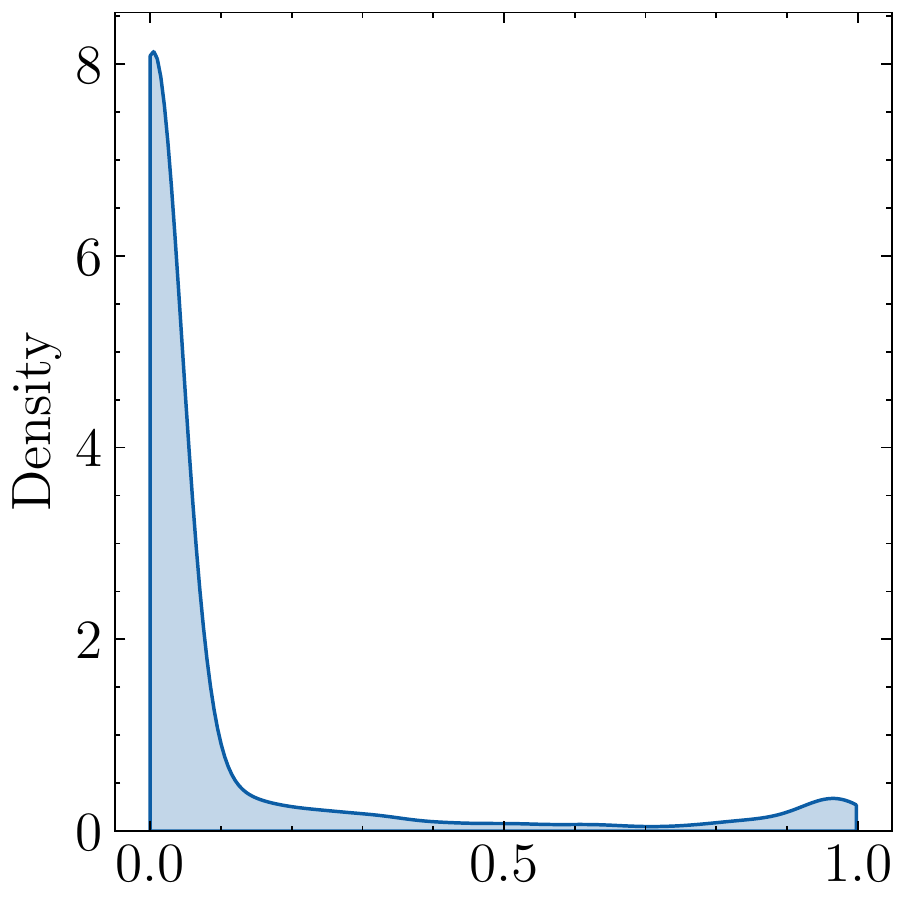}
    \caption{$\gamma = 4$}
\end{subfigure}
\hfill
\begin{subfigure}[b]{0.18\linewidth}
    \centering
    \includegraphics[width=\linewidth]{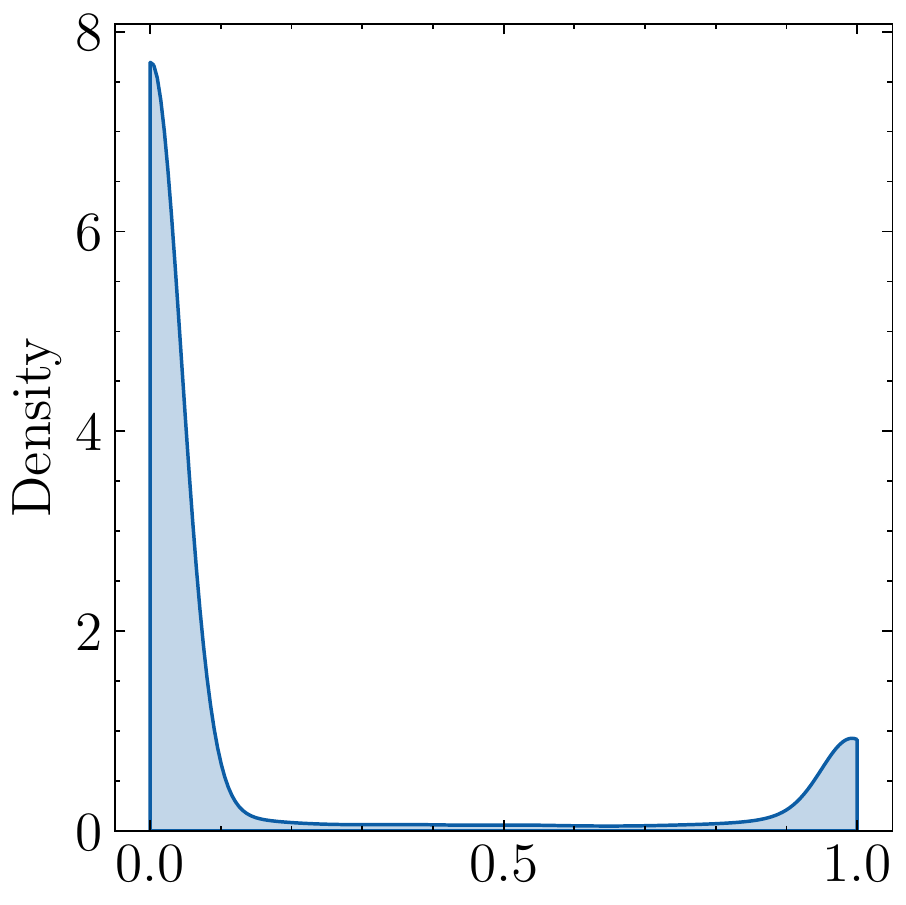}
    \caption{$\gamma = 6$}
\end{subfigure}
\hfill
\begin{subfigure}[b]{0.18\linewidth}
    \centering
    \includegraphics[width=\linewidth]{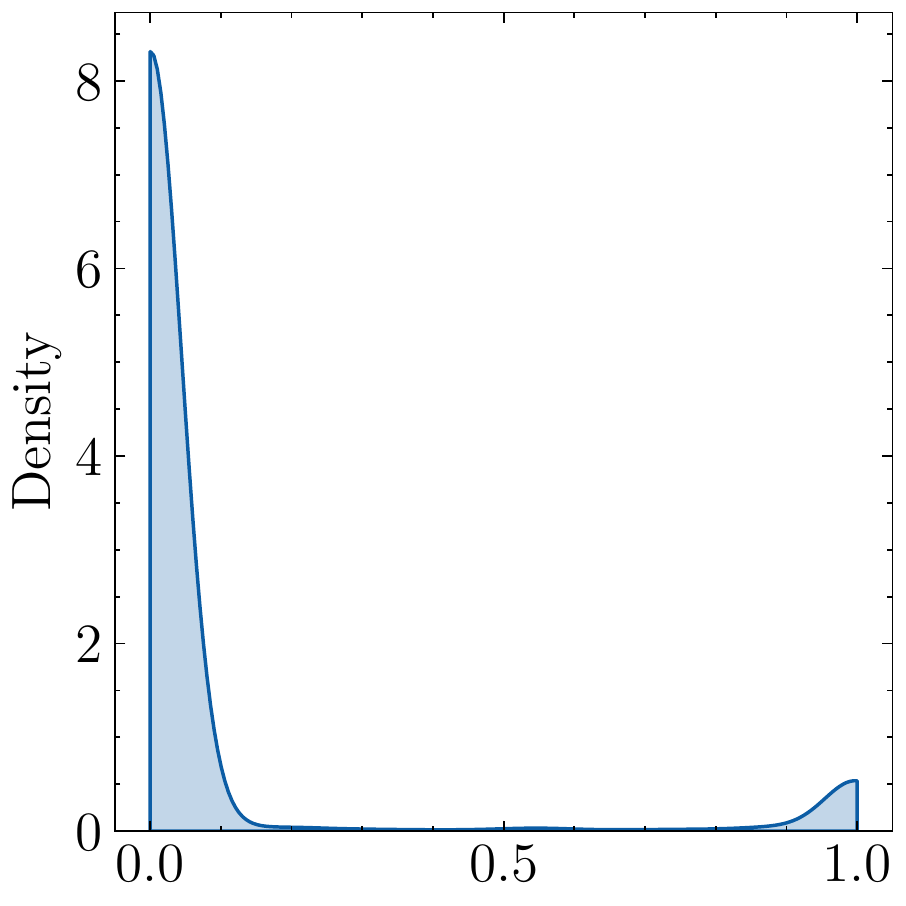}
    \caption{$\gamma = 8$}
\end{subfigure}
\caption{\textbf{Informative sampling -- intensity distribution.} We show the distribution of intensities $\lambda(t)$ over all patients for different levels of informativeness $\gamma$. At $\gamma = 0$ (not shown), all intensities are equal $\lambda_i(t) = 0.5$.}
\label{fig:informative_intensities}
\end{figure}

\begin{figure}[h]
\centering
\begin{subfigure}[b]{0.18\linewidth}
    \centering
    \includegraphics[width=\linewidth]{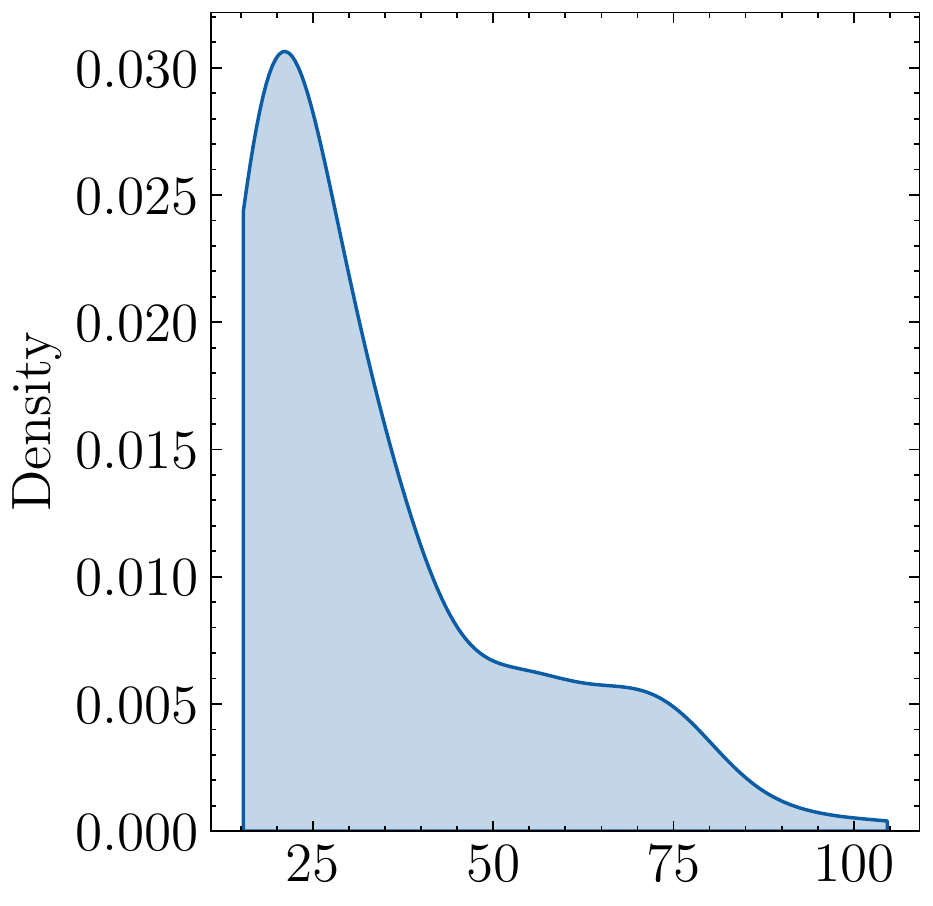}
    \caption{$\gamma = 2$}
\end{subfigure}
\hfill
\begin{subfigure}[b]{0.18\linewidth}
    \centering
    \includegraphics[width=\linewidth]{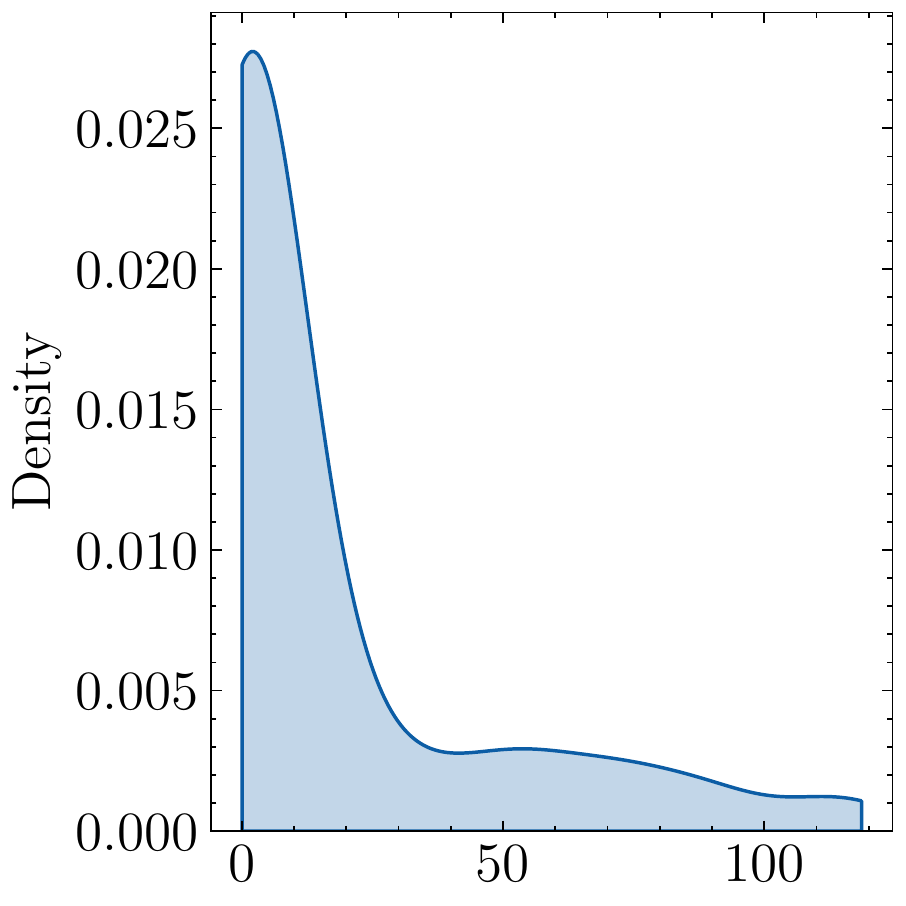}
    \caption{$\gamma = 4$}
\end{subfigure}
\hfill
\begin{subfigure}[b]{0.18\linewidth}
    \centering
    \includegraphics[width=\linewidth]{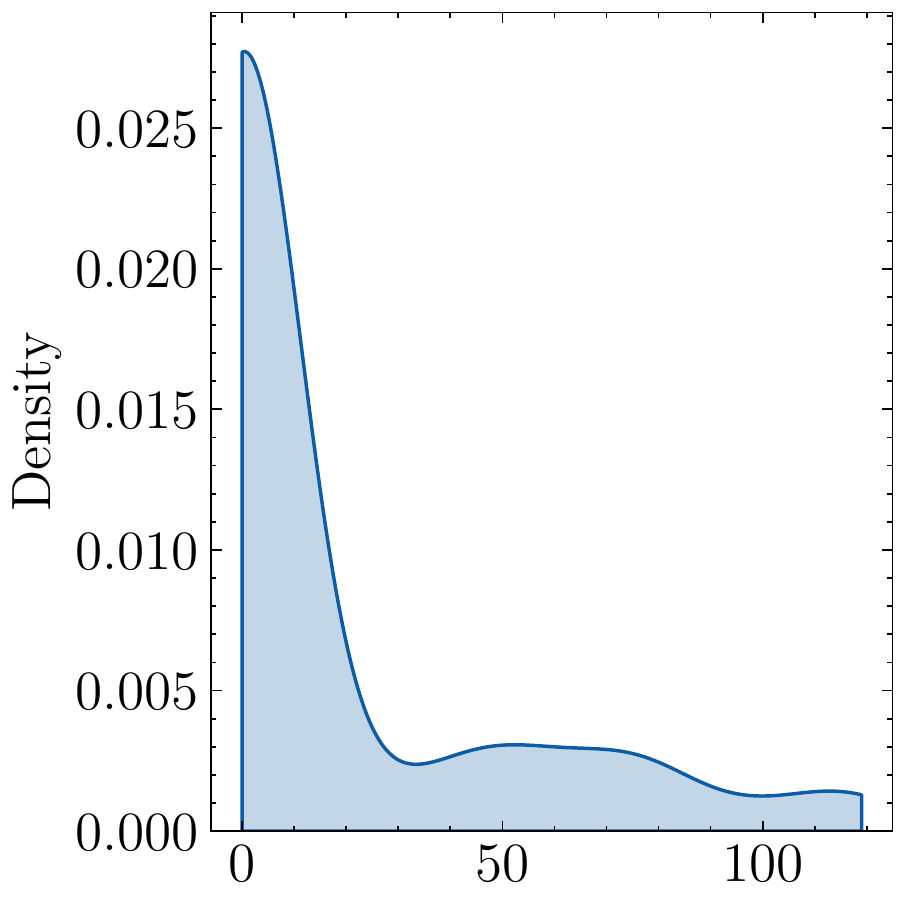}
    \caption{$\gamma = 6$}
\end{subfigure}
\hfill
\begin{subfigure}[b]{0.18\linewidth}
    \centering
    \includegraphics[width=\linewidth]{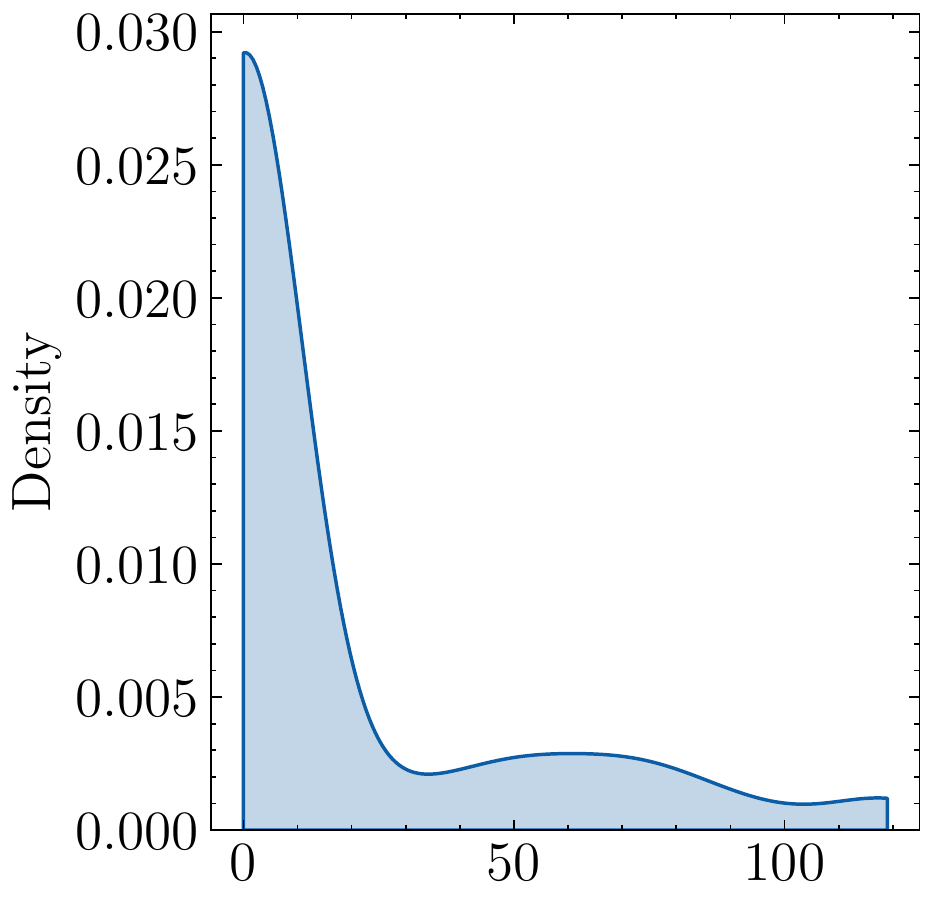}
    \caption{$\gamma = 8$}
\end{subfigure}
\caption{\textbf{Informative sampling -- expected observations.} We show the expected observations over the entire time period considered over all patients for different levels of informativeness $\gamma$. At $\gamma = 0$ (not shown), all intensities are equal $\lambda_i(t) = 0.5$ and all patients have $60$ expected observations.}
\label{fig:informative_expected_obs}
\end{figure}

\begin{figure}[h]
\centering
\begin{subfigure}[b]{0.18\linewidth}
    \centering
    \includegraphics[width=\linewidth]{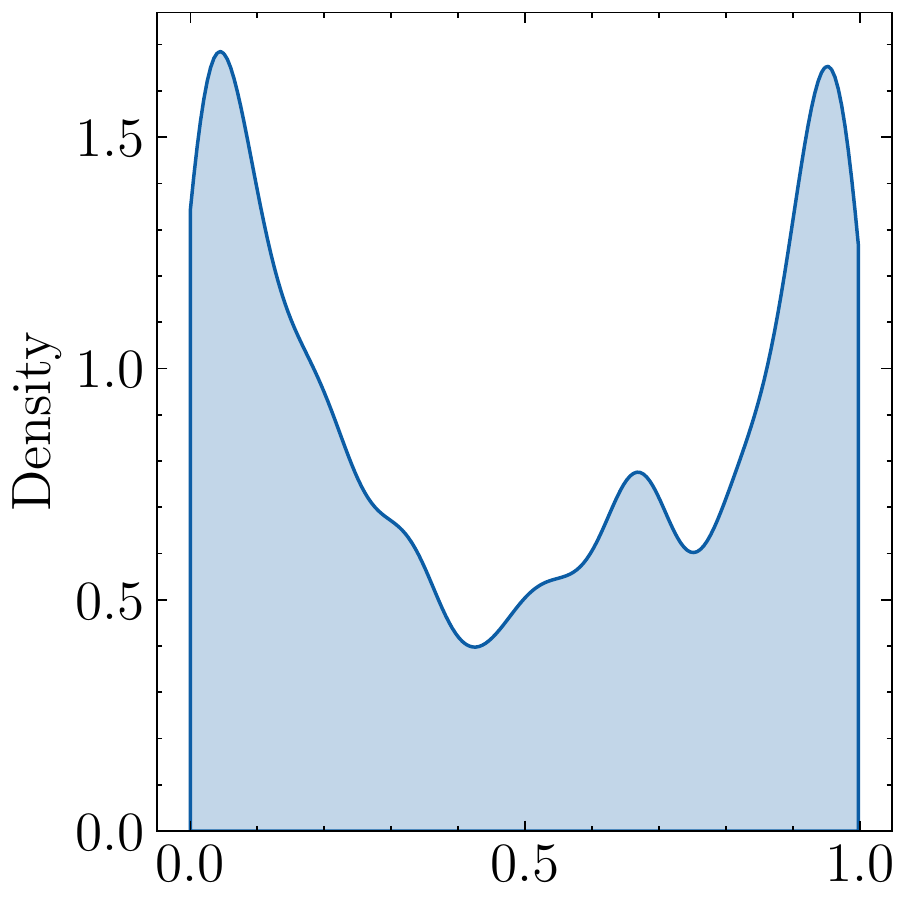}
    \caption{$\gamma = 2$}
\end{subfigure}
\hfill
\begin{subfigure}[b]{0.18\linewidth}
    \centering
    \includegraphics[width=\linewidth]{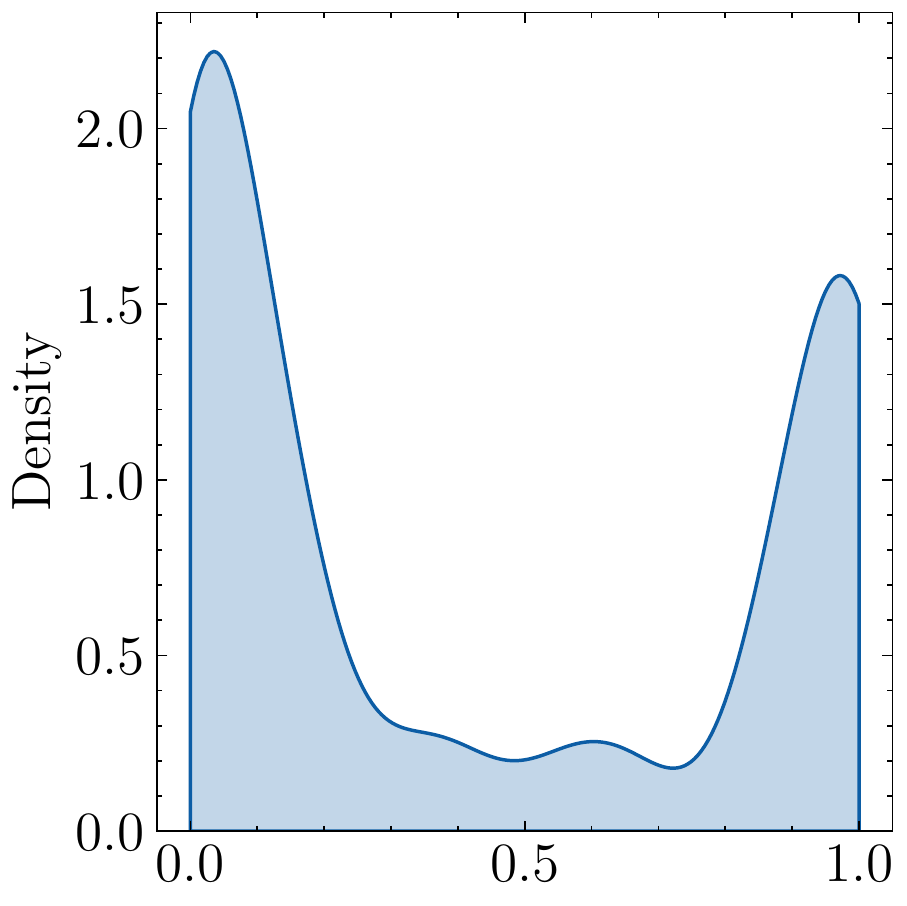}
    \caption{$\gamma = 4$}
\end{subfigure}
\hfill
\begin{subfigure}[b]{0.18\linewidth}
    \centering
    \includegraphics[width=\linewidth]{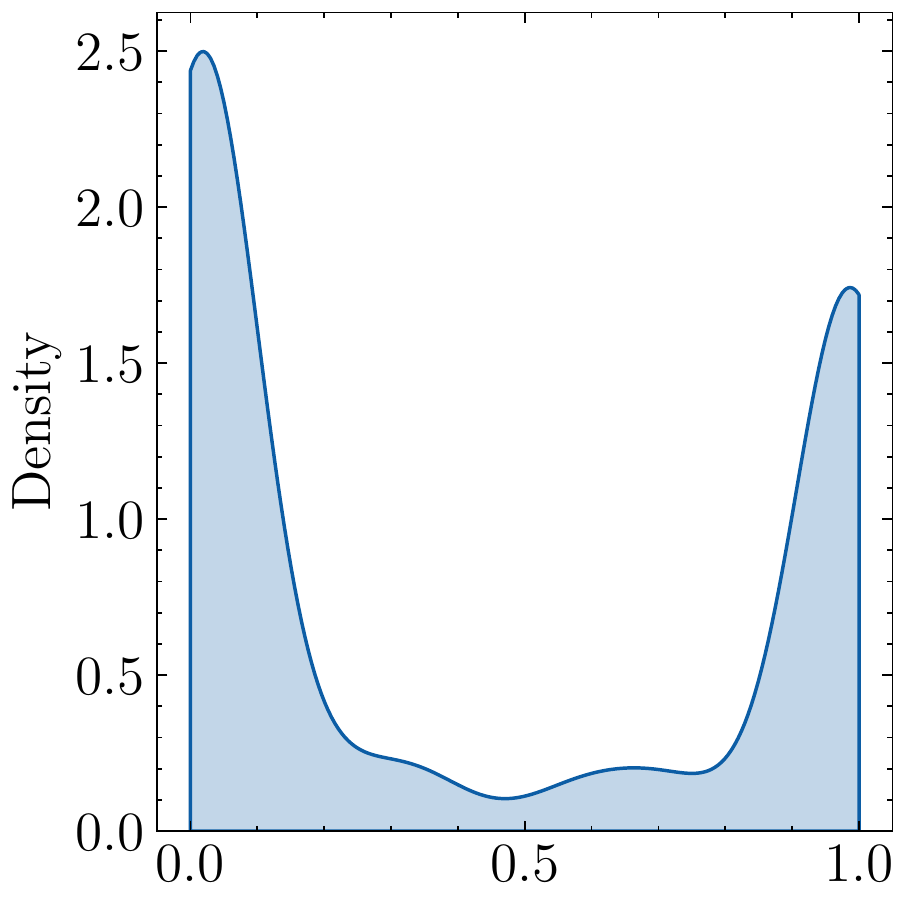}
    \caption{$\gamma = 6$}
\end{subfigure}
\hfill
\begin{subfigure}[b]{0.18\linewidth}
    \centering
    \includegraphics[width=\linewidth]{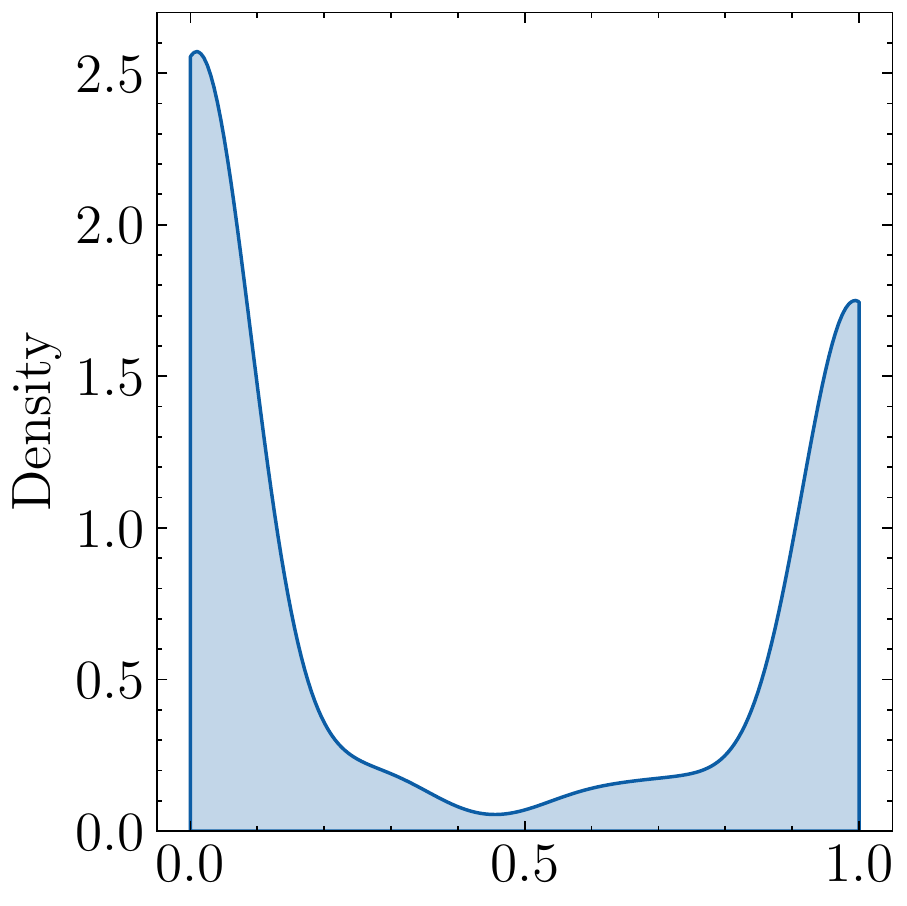}
    \caption{$\gamma = 8$}
\end{subfigure}
\caption{\textbf{Uninformative sampling -- intensity distribution.} We show the distribution of intensities $\lambda(t)$ over all patients for different levels of ``informativeness'' $\gamma$. At $\gamma = 0$ (not shown), all intensities are equal $\lambda_i(t) = 0.5$.}
\label{fig:uninformative_intensities}
\end{figure}

We consider two types of treatment plans: a sequential and a concurrent plan \citep{furuse1999phase, auperin2010meta, curran2011sequential}. We show simulated tumor paths and intensities for several patients in \cref{fig:simulation}. For the informative observation process \cref{eq:obs_process}, we visualize the intensity distribution at different levels of $\gamma$ in \cref{fig:informative_intensities,fig:informative_expected_obs}. For the uninformative observation process, we show the intensity distributions in \cref{fig:uninformative_intensities}.

For all experiments, we generate patient trajectories over 120 days. For training, we split the data for forecasting to have a lookback window of seven days and a maximum forecasting horizon of five days.

\clearpage
\section{Additional Results}
\label{app:Additional_results}

In this section, we present additional results to further validate the proposed TESAR-CDE. First, we evaluate the predicted observation intensities. Second, we analyze the sensitivity of the multitask model to hyperparameter $\alpha$.

\begin{figure*}[h]
\centering
\includegraphics[width=0.55\linewidth, trim={0 0 0 0}]{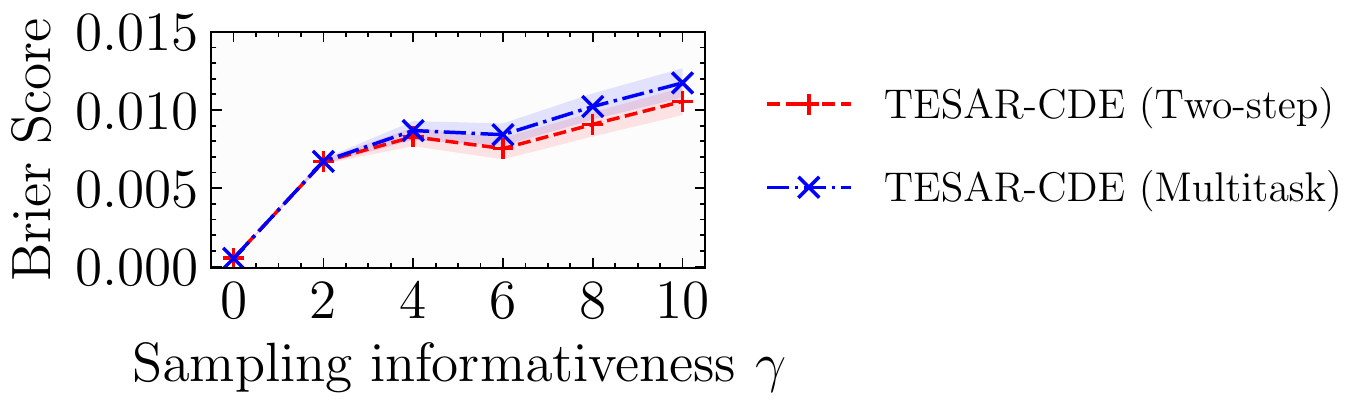}
\caption{\textbf{Evaluating the intensity prediction at varying informativeness $\gamma$.} We show the Brier Score $\pm$ SE (lower is better) over ten runs at increasing levels of informativeness $\gamma$, keeping the forecasting horizon $\tau = 1$.}
\label{fig:results_intensity_prediction}
\end{figure*}

We evaluate how accurate TESAR-CDE predicts the observation intensities in terms of the Brier Score: $\text{BS} = \sum_{t=0}^T \! \sum_{\tau=0}^{\tau_\text{max}} \left( \lambda_i(t+\tau) - \hat{\lambda}_{i, t}(t+\tau) \right)^2$, see \cref{fig:results_intensity_prediction}. Generally, both versions of our method can learn to accurately predict the observation intensities, with the two-step TESAR-CDE performing slightly better than the multitask configuration. These findings are consistent with our motivation of the multitask setup: while the two-step model learns a (generally better) model of the intensity itself using all available information, these more accurate intensities do not help with potential outcome prediction. Additionally, we find that the Brier score increases with informativeness for both models. This trend indicates that more informativeness makes it harder to learn to predict the observation intensity. We hypothesize that this is due to observing in general becoming more rare as increases.

\begin{figure*}[th]
\centering
\includegraphics[width=0.35\linewidth, trim={0 0 0 0}]{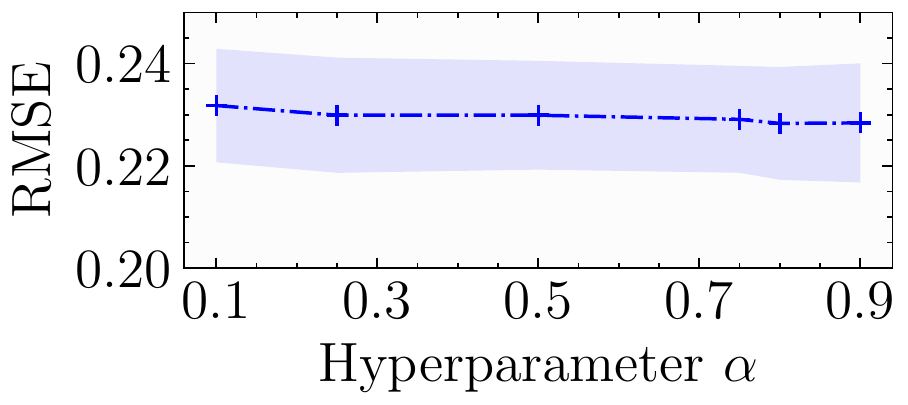}
\caption{\textbf{Evaluating the multitask configuration's outcome prediction for different levels of hyperparameter $\alpha$.} We show the RMSE $\pm$ SE over ten runs, while keeping the level of informativeness fixed at $\gamma = 6$ and averaging over $\tau \in \{1, \dots, 5\}$.}
\label{fig:results_alpha_sensitivity}
\end{figure*}

Next, we evaluate performance for the multitask TESAR-CDE for different values of its hyperparameter $\alpha$, see \cref{fig:results_alpha_sensitivity}. As the shared representation is only trained for outcome prediction, the only point of having the hyperparameter is to scale both terms such that they roughly influence the early stopping in the same way. We see that our method is robust to different values of this hyperparameter and that scaling them to approximately the same magnitude (using $0.8$) results in good performance in practice. (Note that the loss terms are weighted by $\alpha$ and $(1-\alpha)$, which is why we restrict to be strictly between $0$ and $1$.)

\end{document}